\documentclass[11pt]{article}
\pdfoutput=1

\usepackage{fullpage}
\usepackage{amsmath}
\usepackage{amssymb}
\usepackage{graphicx} 

\usepackage{fancybox,graphicx}
\usepackage[T1]{fontenc}
\usepackage[latin1]{inputenc}
\usepackage{float}

\floatstyle{ruled}
\newfloat{alg}{tbp}{loa}
\floatname{alg}{Algorithm}

{
\rm
\begin{tabbing}
....\=...\=...\=...\=...\=  \+ \kill
} %
{\end{tabbing}
}

\newtheorem{definition}{Definition}[section]
\newtheorem{lemma}{Lemma}[section]

\newtheorem{theorem}{Theorem}[section]
\newtheorem{proposition}{Proposition}[section]

\newcommand{\cM}{{\cal M}}
\newcommand{\cN}{{\cal N}}
\newcommand{\Real}{{\mathbb R}}
\newcommand{\rE}{{\mathbb E}}
\newcommand{\cF}{{\cal F}}

\newcommand{\BlackBox}{\rule{1.5ex}{1.5ex}}

\begin{document}

\title{High Dimensional Nonlinear Learning using Local Coordinate Coding}
\date{}

\author{Kai Yu \\ NEC Laboratories America \\ kyu@sv.nec-labs.com \\
\and Tong Zhang \\ Rutgers University \\ tzhang@stat.rutgers.edu
}

\maketitle

\begin{abstract}
This paper introduces a new method for semi-supervised learning on high
dimensional nonlinear manifolds, which includes a phase
of unsupervised basis learning and a phase of supervised function
learning. The learned bases provide a set of anchor points to
form a local coordinate system, such that each data point $x$ on the
manifold can be locally approximated by a linear combination of its
nearby anchor points, with the linear weights offering a
local-coordinate coding of $x$. We show that a high dimensional
nonlinear function
can be approximated by a global linear function with respect to this
coding scheme, and the approximation quality is ensured by the
locality of such coding. The method turns a difficult
nonlinear learning problem into a simple global linear learning problem, 
which 
overcomes some drawbacks of traditional local learning methods. The
work also gives a theoretical justification to the empirical success
of some biologically-inspired models using sparse coding of sensory
data, since a local coding scheme must be sufficiently sparse.
However, sparsity does not always satisfy locality conditions, and
can thus possibly lead to suboptimal results. The properties and
performances of the method are empirically verified on synthetic
data, handwritten digit classification, and object recognition
tasks.
\end{abstract}

\section{Introduction}

Consider the problem of learning a nonlinear function $f(x)$ in
high dimension: $x \in \Real^d$ with large $d$. We are given a set of
labeled data $(x_1,y_1),\ldots,(x_n,y_n)$ drawn from an unknown
underlying distribution. Moreover, assume that we observe a set of
unlabeled data $x \in \Real^d$ from the same distribution.
If the dimensionality $d$ is large compared to $n$,
then the traditional statistical theory predicts over-fitting due to the
so called ``curse of dimensionality''.
One intuitive
argument for this effect is that when the dimensionality becomes larger,
pairwise distances between two similar data points become larger as well.
Therefore one needs more data points to adequately fill in the empty space.
However, for many real problems with high dimensional data, we do not observe
this so-called curse of dimensionality.
This is because although data are physically represented
in a high-dimensional space, they often (approximately) lie on a manifold
which has a much smaller intrinsic dimensionality.

This paper proposes a new method that can take advantage of the
manifold geometric structure to learn a nonlinear function in high dimension.
The main idea is to locally embed points on the manifold into a
lower dimensional space, expressed as coordinates with respect to a
set of anchor points. Our main observation is simple but very
important: we show that a nonlinear function on the manifold can be
effectively approximated by a linear function with such a coding
under appropriate localization conditions. Therefore by using {\em Local
Coordinate Coding}, we turn a very difficult high dimensional
nonlinear learning problem into a much simpler linear learning
problem, which has been extensively studied in the literature. This
idea may also be considered as a high dimensional generalization of
low dimensional local smoothing methods in the traditional
statistical literature.

\section{Local Coordinate Coding}

We are interested in learning a smooth function $f(x)$ defined on
a high dimensional space $\Real^d$.
Let $\|\cdot\|$ be a norm on $\Real^d$. Although we do not restrict
to any specific norm, in practice, one often employs the Euclidean
norm (2-norm): $\|x\| = \|x\|_2=\sqrt{x_1^2+\cdots+x_d^2}$.
\begin{definition}[Lipschitz Smoothness]
  A function $f(x)$ on $\Real^d$ is
  $(\alpha,\beta,p)$-Lipschitz
  smooth with respect to a norm $\|\cdot\|$ if
  \[
  |f(x')-f(x)| \leq \alpha \|x-x'\| ,
  \]
  and
  \[
  |f(x')-f(x) - \nabla f(x)^\top (x'-x)| \leq \beta \|x-x'\|^{1+p} ,
  \]
  where we assume $\alpha,\beta>0$ and $p \in (0,1]$.
\end{definition}

Note that if the Hessian of $f(x)$ exists, then we may take $p=1$.
Learning an arbitrary Lipschitz smooth function on $\Real^d$ can be
difficult due to the curse of dimensionality. That is, the number of
samples required to characterize such a function $f(x)$
can be exponential in $d$.
However, in many practical applications, one often observes that
the data we are interested in lie approximately on a
manifold $\cM$ which is embedded into $\Real^d$. Although $d$ is large,
the intrinsic dimensionality of $\cM$ can be much smaller.
Therefore if we are only interested in learning $f(x)$ on $\cM$,
then the complexity should depend on the intrinsic dimensionality of
$\cM$ instead of $d$.

In this paper, we approach this problem by
introducing the idea of localized coordinate coding. The formal
definition of (non-localized) coordinate coding is given below,
where we represent a point in $\Real^d$
by a linear combination of a set of ``anchor points''.
Later we show it is sufficient to choose a set of ``anchor points''
with cardinality depending on the intrinsic dimensionality
of the manifold rather than $d$.
\begin{definition}[Coordinate Coding]
  A coordinate coding is a pair $(\gamma,C)$,
  where $C \subset \Real^d$ is a set of anchor points,
  and $\gamma$ is a map of $x \in \Real^d$ to
  $[\gamma_v(x)]_{v \in C} \in R^{|C|}$ such that
  $\sum_{v} \gamma_v(x) =1$. It induces
  the following physical approximation of $x$ in $\Real^d$:
  \[
  \gamma(x)=\sum_{v \in C} \gamma_v(x) v  .
  \]
  Moreover, for all $x \in \Real^d$, we define the coding norm as
  \[
  \|x\|_\gamma= \left(\sum_{v \in C} \gamma_v(x)^2\right)^{1/2} .
  \]
\end{definition}

The quantity $\|x\|_\gamma$ will become useful in our learning
theory analysis. The condition $\sum_{v} \gamma_v(x) =1$ follows
from the shift-invariance requirement, which means that the coding
should remain the same if we use a different origin of the $\Real^d$
coordinate system for representing data points.
However, if in practice we can find a good origin for
the global coordinate system in $\Real^d$, and if all points on $\cM$
are close to it, then the shift-invariance requirement may become 
less important.
\begin{proposition} \label{prop:shift-inv}
  The map $x \to \sum_{v \in C} \gamma_v(x) v$ is invariant
  under any shift of the origin for representing data
  points in $\Real^d$ if and only if $\sum_{v} \gamma_v(x) =1$.
\end{proposition}

The importance of the coordinate coding concept is that if a
coordinate coding is sufficiently localized, then a nonlinear
function can be approximate by a linear function with respect to the
coding. This critical observation, illustrate in the following
linearization lemma, is the foundation of our approach.
\begin{lemma}[Linearization] \label{lem:coding}
  Let $(\gamma,C)$ be an arbitrary coordinate coding on $\Real^d$.
  Let $f$ be an $(\alpha,\beta,p)$-Lipschitz smooth function.
  We have for all $x \in \Real^d$:
  \[
  \left|f(x) - \sum_{v \in C} \gamma_v(x) f(v)\right|
  \leq \alpha \left\|x- \gamma(x)\right\|
  + \beta \sum_{v \in C} |\gamma_v(x)|  \left\|v- \gamma(x)\right\|^{1+p} .
  \]
\end{lemma}

To understand this result, we note that on the left hand side, a
nonlinear function $f(x)$ in $R^d$ is approximated by a linear
function $\sum_{v \in C} \gamma_v(x) f(v)$ with respect to the
coding $\gamma(x)$, where $[f(v)]_{v \in C}$ is the set of
coefficients to be estimated from data. The quality of this
approximation is bounded by the right hand side, which has two
terms: the first term $\|x - \gamma(x)\|$ means $x$ should be close
to its physical approximation $\gamma(x)$, and the second term means
that the coding should be localized. The quality of a coding
$\gamma$ with respect to $C$ can be measured by the right hand side.
For convenience, we introduce the following definition, which
measures the locality of a coding.
\begin{definition}[Localization Measure]
  Given $\alpha,\beta,p$, and coding $(\gamma,C)$, we define
\[
Q_{\alpha,\beta,p}(\gamma,C)= \rE_{x}
\left[ \alpha  \|x - \gamma(x)\| + \beta
\sum_{v \in C} |\gamma_v(x)|  \left\|v- \gamma(x)\right\|^{1+p} \right] .
\]
\end{definition}
Observe that in $Q_{\alpha,\beta,p}$,
$\alpha,\beta,p$ may be regarded as tuning parameters;
we may also simply pick $\alpha=\beta=p=1$.
Since the quality function $Q_{\alpha,\beta,p}(\gamma,C)$ only depends on unlabeled data, in principle,
we can find $[\gamma,C]$ by optimizing this quality using unlabeled data.
Later, we will consider simplifications of this objective function that
are easier to compute.

Next we show that if the data lie on a manifold, then the complexity
of local coordinate coding depends on the intrinsic manifold
dimensionality instead of $d$. We first define manifold and its
intrinsic dimensionality.
\begin{definition}[Manifold] \label{def:manifold}
  A subset $\cM \subset \Real^d$ is called a $p$-smooth ($p >0$)
  manifold with intrinsic dimensionality $m=m(\cM)$
  if there exists a constant
  $c_p(\cM)$ such that given any $x \in \cM$, there exists $m$ vectors
  $v_1(x),\ldots,v_{m}(x) \in \Real^d$ so that
  $\forall x' \in \cM$:
  \[
  \inf_{\gamma \in \Real^{m}} \left\|x'-x - \sum_{j=1}^m \gamma_j v_j(x)\right\| \leq c_p(\cM) \|x' - x\|^{1+p} .
  \]
\end{definition}
This definition is quite intuitive. The smooth manifold structure
implies that one can approximate a point in $\cM$ effectively using
local coordinate coding. Note that for a typical manifold with
well-defined curvature, we can take $p=1$.

\begin{definition}[Covering Number]  \label{def:covering}
Given any subset $\cM \subset \Real^d$, and $\epsilon>0$.
The covering number, denoted as $\cN(\epsilon,\cM)$, is the
smallest cardinality of an $\epsilon$-cover $C\subset \cM$.
That is,
  \[
  \sup_{x \in \cM} \inf_{v \in C} \|x - v\| \leq \epsilon .
  \]
\end{definition}
For a compact manifold with intrinsic dimensionality $m$, there exists
a constant $c(\cM)$ such that its covering number is bounded by
\[
\cN(\epsilon,\cM) \leq c(\cM) \epsilon^{-m} .
\]
The usual statistical definition of dimensionality only involves
the covering number $|C|$.
However, the manifold intrinsic dimensionality is also important 
by itself in our analysis.

The following result shows that there exists a local coordinate
coding to a set of anchor points $C$ of cardinality $O(m(\cM)
\cN(\epsilon,\cM))$ such that any $(\alpha,\beta,p)$-Lipschitz
smooth function can be linearly approximated using local coordinate
coding up to the accuracy $O(\sqrt{m(\cM)}\epsilon^{1+p})$.

\begin{theorem}[Manifold Coding] \label{thm:manifold-coding}
  If the data points $x$
  lie on a compact $p$-smooth manifold $\cM$, and the norm is defined
  as $\|x\| = (x^\top A x)^{1/2}$ for some positive definite matrix $A$.
  Then given any $\epsilon>0$, there exist anchor points $C \subset \cM$
  and coding $\gamma$ such that
  \begin{align*}
    & |C| \leq (1+ m) \cN(\epsilon,\cM) , \\
    & Q_{\alpha,\beta,p}(\gamma,C) \leq [\alpha c_p(\cM)
    + (1+ \sqrt{m} + 2^{1+p} \sqrt{m})\beta] \, \epsilon^{1+p} ,
  \end{align*}
  where $m=m(\cM)$.
  Moreover, for all $x \in \cM$, we have
  $\|x\|_\gamma^2 \leq 1 + (1+\sqrt{m})^2$.
\end{theorem}

The approximation result in Theorem~\ref{thm:manifold-coding} means that
the complexity of linearization in Lemma~\ref{lem:coding} depends only
on the intrinsic dimension $m(\cM)$ of $\cM$ instead of $d$.
Although this result is proved for manifolds, it is important to
observe that the coordinate coding method proposed in this paper
does not require the data to lie precisely on a manifold, and it
does not require knowing $m(\cM)$. In fact, similar results hold
even when the data only approximately lie on a manifold.

In the next section, we characterize the learning complexity of the
local coordinate coding method. It implies that linear prediction
methods can be used to effectively learn nonlinear functions on a
manifold. The nonlinearity is fully captured by the coordinate
coding map $\gamma$ (which can be a nonlinear function).
This approach has some great advantages because the problem of learning
local-coordinate coding 
is much simpler than direct nonlinear learning:
\begin{itemize}
\item Learning $(\gamma,C)$ only requires unlabeled
  data, and the number of unlabeled data
  can be significantly more than the number of labeled data.
  This also prevents overfitting with respect to labeled data.
\item In practice, we do not have to find the optimal coding because
  the coordinates are merely features for linear supervised
  learning. This significantly simplifies the optimization problem.
  Consequently, it is more robust than standard approaches to
  nonlinear learning that direct optimize nonlinear functions on labeled
  data (e.g., neural networks).
\end{itemize}

\section{Learning Theory}

In machine learning, we minimize the expected loss
$\phi(f(x),y)$ with respect to the underlying distribution
\[
\rE_{x,y} \phi(f(x),y)
\]
within a function class $f(x) \in \cF$.
In this paper, we are interested in the function class
\[
\cF_{\alpha,\beta,p} =
\{f(x): (\alpha,\beta,p)-\text{Lipschitz smooth function in } \Real^d \} .
\]

The local coordinate coding method considers a linear approximation
of functions in $\cF_{\alpha,\beta,p}$ on the data manifold. Given a
local-coordinate coding scheme $(\gamma,C)$, we approximate each
$f(x) \in \cF_{\alpha,\beta,p}^a$ by
\[
f(x) \approx f_{\gamma,C}(\hat{w},x)=\sum_{v \in C} \hat{w}_v \gamma_v(x) ,
\]
where we estimate the coefficients using ridge regression as:
\begin{equation}
[\hat{w}_v] = \arg\min_{[w_v]}
\left[
\sum_{i=1}^n
\phi\left(f_{\gamma,C}(w, x_i),y_i\right) + \lambda
\sum_{v \in C} w_v^2
\right] . \label{eq:est}
\end{equation}

Given a loss function $\phi(p,y)$,
let $\phi_1'(p,y)=\partial \phi(p,y)/\partial p$.
For simplicity, in this paper we only consider convex
Lipschitz loss function, where $|\phi_1'(p,y)| \leq B$.
This includes the standard classification loss functions such as
logistic regression and SVM (hinge loss), both with $B=1$.
\begin{theorem}[Generalization Bound] \label{thm:gen}
  Suppose $\phi(p,y)$ is Lipschitz: $|\phi_1'(p,y)| \leq B$.
  Consider coordinate coding $(\gamma,C)$, and
  the estimation method (\ref{eq:est}) with random training
  examples $S_n=\{(x_1,y_1),\ldots,(x_n,y_n)\}$. Then
  the expected generalization error satisfies the inequality:
  \begin{align*}
    & \rE_{S_n} \; \rE_{x,y} \phi(f_{\gamma,C}(\hat{w},x),y) \\
  \leq& \inf_{f \in \cF_{\alpha,\beta,p}}
  \left[\rE_{x,y} \phi\left(f(x),y\right) + \lambda
    \sum_{v \in C} f(v)^2 \right]
  + \frac{B^2}{2 \lambda n} \rE_x \|x\|_\gamma^2
  + B Q_{\alpha,\beta,p}(\gamma,C) .
\end{align*}
\end{theorem}

If we choose the regularization parameter $\lambda$ that optimizes the bound,
then the right hand side becomes
\begin{equation}
\inf_{f \in \cF_{\alpha,\beta,p}}
\left[\rE_{x,y} \phi\left(f(x),y\right) + B
  \sqrt{\frac{2}{n}\sum_{v \in C} f(v)^2 \; \rE_x \|x\|_\gamma^2} \right]
+ B Q_{\alpha,\beta,p}(\gamma,C) . \label{eq:gen}
\end{equation}
In particular, if we find $(\gamma,C)$ at
some $\epsilon>0$, then Theorem~\ref{thm:manifold-coding} implies
the following simplification for any $f \in \cF_{\alpha,\beta,p}$
such that $|f(x)| \leq A$ for a fixed constant $A$, then the bound
on the generalization error becomes:
\[
\rE_{x,y} \phi\left(f(x),y\right) +
O \left[\sqrt{\epsilon^{-m(\cM)}/n}
+ \epsilon^{1+p} \right] .
\]
By optimizing over $\epsilon$, we obtain a bound:
$\rE_{x,y} \phi\left(f(x),y\right) +
O (n^{-(1+p)/(2+2p+m(\cM))})$.

By combining Theorem~\ref{thm:manifold-coding} and
Theorem~\ref{thm:gen}, we can immediately obtain the following
simple consistency result. It shows that the algorithm can learn an
arbitrary nonlinear function on manifold when $n \to \infty$. Note
that Theorem~\ref{thm:manifold-coding} implies that the convergence
only depends on the intrinsic dimensionality of the manifold $\cM$,
not $d$.
\begin{theorem}[Consistency] \label{thm:consistency}
  Suppose the data lie on a compact manifold $\cM \subset \Real^d$,
  and the norm $\|\cdot\|$ is the Euclidean norm in $\Real^d$.
  If loss function $\phi(p,y)$ is Lipschitz. As $n \to \infty$,
  we choose $\alpha,\beta \to \infty$, $\alpha/n,\beta/n \to 0$
  ($\alpha,\beta$ depends on $n$), and $p=0$.
  Then it is possible to 
  find coding $(\gamma,C)$ using unlabeled data
  such that $|C|/n \to 0$ and
  $Q_{\alpha,\beta,p}(\gamma,C) \to 0$.
  If we pick  $\lambda n \to \infty$, and $\lambda |C| \to 0$.
  Then the local coordinate coding method (\ref{eq:est})
  is consistent as $n \to \infty$:
  \[
  \lim_{n \to \infty} \rE_{S_n} \; \rE_{x,y} \phi(f(\hat{w},x),y)
  =\inf_{f: \cM \to \Real}
  \rE_{x,y} \phi\left(f(x),y\right) .
  \]
\end{theorem}

\section{Practical Learning of Coding}

Given a coordinate coding $(\gamma,C)$, we can use (\ref{eq:est}) to
learn a nonlinear function in $\Real^d$. We showed that $(\gamma,C)$
can be obtained by optimizing $Q_{\alpha,\beta,p}(\gamma,C)$. In
practice, we may also consider the following simplifications of the
localization term: 
\[
\sum_{v \in C} |\gamma_v(x)|  \left\|v- \gamma(x)\right\|^{1+p}
\approx
\sum_{v \in C} |\gamma_v(x)|  \left\|v- x\right\|^{1+p} .
\]
Note that we may simply chose $p=0$ or $p=1$. The formulation is related
to sparse coding \cite{raina:self-taught}
which has no locality constraints with $p=-1$.
In this representation, we may either enforce
the constraint $\sum_v \gamma_v(x)=1$ or remove it for simplicity
(in such case, we assume that the coordinate
origin is appropriately chosen so that
the shift-invariance requirement is not important).
Putting the above together, we try to optimize the following
objective function in practice:
\[
Q(\gamma,C) =
\rE_x
\inf_{[\gamma_v]}
\left[
\left\|x  - \sum_{v \in C} \gamma_v v\right\|^2
+ \mu \sum_{v \in C} |\gamma_v| \|v - x\|^{2}
\right] .
\]

\section{Relationship to Other Methods}

Our work is related to several existing approaches in the literature
of machine learning and statistics. The first class of them is
nonlinear manifold learning, such as LLE \cite{lle}, Isomap
\cite{isomap}, and Laplacian Eigenmaps \cite{lap_eig}. These methods
find {\em global} coordinates of data manifold based on a pre-computed
affinity graph of data points. The use of affinity graphs requires
expensive computation and lacks a coherent way of generalization to
new data. Our method learns a compact set of bases to form local
coordinates, which has a linear complexity with respect to data size
and can naturally handle unseen data. More importantly, local
coordinate coding has a direct connection to function approximation
on manifold, and thus provides a sound unsupervised pre-training
method to facilitate further supervised learning tasks.

Another set of related models are local models in statistics, such
as local kernel smoothing and local regression, both traditionally
using fixed-bandwidth kernels. Local kernel smoothing can be
regarded as a zero-order method; while local regression is
higher-order, including local linear regression as the 1st-order
case. Traditional local methods are not widely used in machine
learning practice, because data with non-uniform distribution on the
manifold require to use adaptive-bandwidth kernels. The problem can
be somehow alleviated by using $K$-nearest neighbors. However,
adaptive kernel smoothing still suffers from the high-dimensionality
and noise of data. On the other hand, higher-order methods are
computationally expensive and prone to overfitting, because they are
highly flexible in locally fitting many segments of data in
high-dimensional spaces. Our method can be seen as a generalized
1st-order local method with basis learning and adaptive locality.
Compared to local linear regression, the learning is achieved by
fitting a single globally linear function with respect to a set of
learned local coordinates, which is much less prone to overfitting
and computationally much cheaper.
This means that our method achieves
better balance between local and global aspects of learning. The
importance of such balance has been recently discussed in \cite{ZaRi09}.

Finally, local-coordinate coding draws connections to vector
quantization (VQ) coding and sparse coding, which have been widely
applied in processing of sensory data, such as acoustic and image
signals. Learning linear functions of VQ codes can be regarded as a
generalized zero-order local method with adaptive basis learning.
Our method has an intimate relationship with sparse coding. In fact,
we can regard local coordinate coding as locally constrained sparse
coding. Inspired by biological visual systems, people has been
arguing sparse features of signals are useful for learning
\cite{raina:self-taught}. However, to the best of our knowledge,
there is no analysis in the literature that directly answers the
question why sparse codes can help learning nonlinear functions in
high dimensional spaces. Our work reveals an important finding
--- a good first-order approximation to nonlinear function requires
the codes to be local, which consequently requires the codes to be
sparse. However, sparsity does not always guarantee locality
conditions. Our experiments demonstrate that sparse coding is
helpful for learning only when the codes are local. Therefore
locality is more essential for coding, and sparsity is a consequence
of such a condition.

Properties of related methods discussed in this section are
compared in Table~\ref{tab:comparison}.

\begin{table}[htb]
\centering
\begin{tabular}{|c|c|c|c|} \hline
method & dimension &  basis learning & approximation  power \\ \hline \hline
kernel smoothing & low  & no & 0th order \\ \hline
local linear regression& low & no & 1st order \\ \hline
$K$-nearest neighbor & high & no & 0th order \\ \hline
Vector Quantization (VQ) & high & yes  & 0th order \\ \hline
Local Coordinate Coding (LCC)& high  & yes & 1st order \\ \hline
\end{tabular} 
\caption{Comparison of Related Methods} \label{tab:comparison}
\end{table}

\section{Experiments}

We use three experiments to demonstrate various points of the 
theoretical claims. In particular, the importance of coding locality
and the robustness of various methods to high data dimensionality.

\subsection{Synthetic Data}

\begin{figure} \footnotesize
  \centering
  \begin{tabular}{{c}{c}{c}{c}}
  \includegraphics[width=0.23\textwidth]{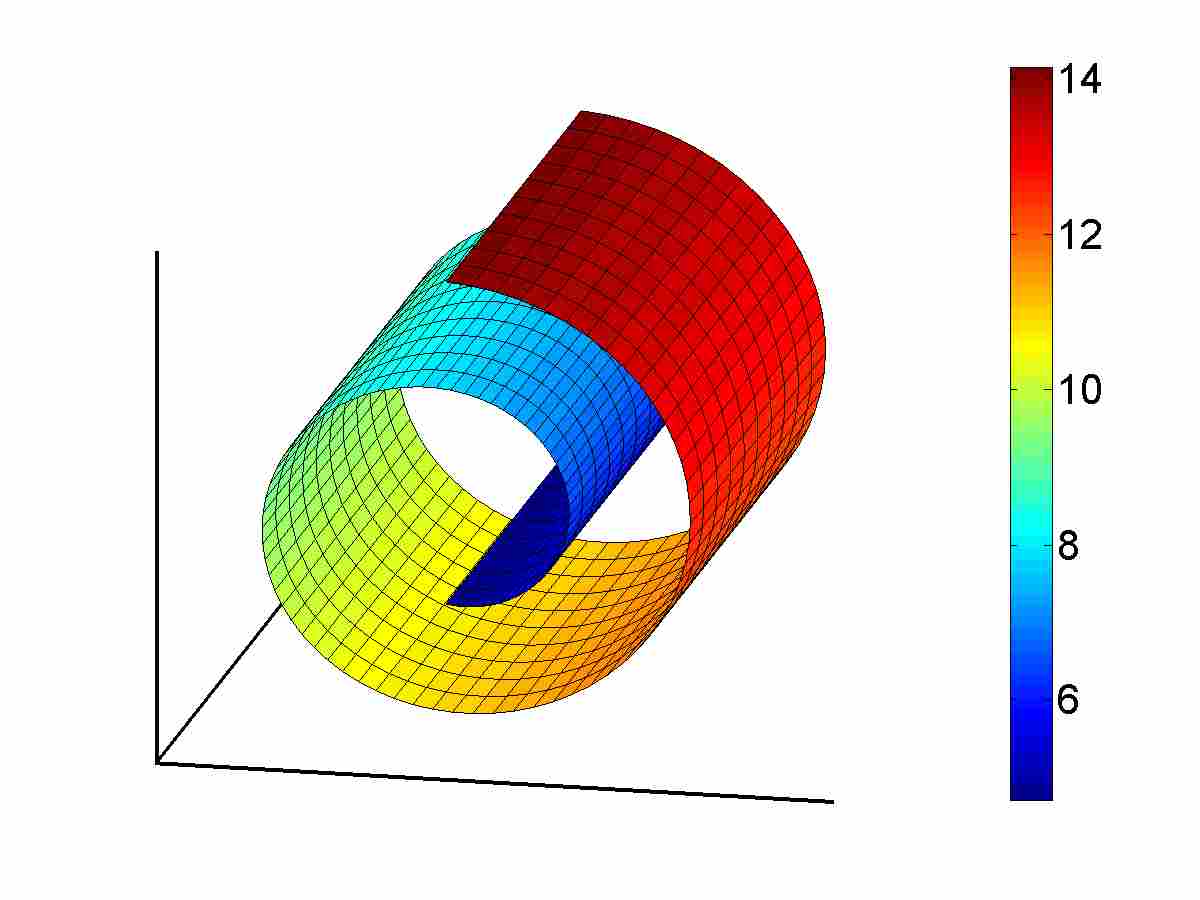} &
\includegraphics[width=0.23\textwidth]{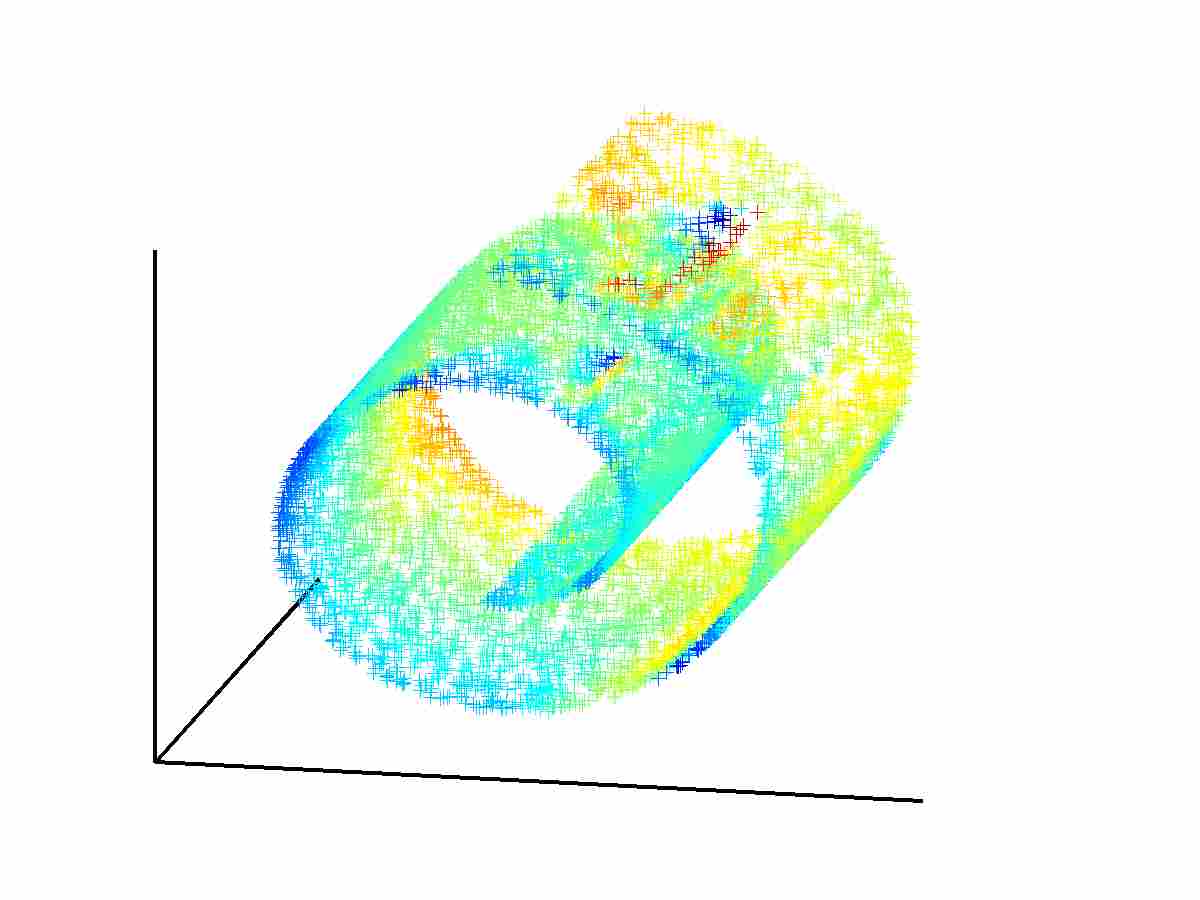} &
\includegraphics[width=0.23\textwidth]{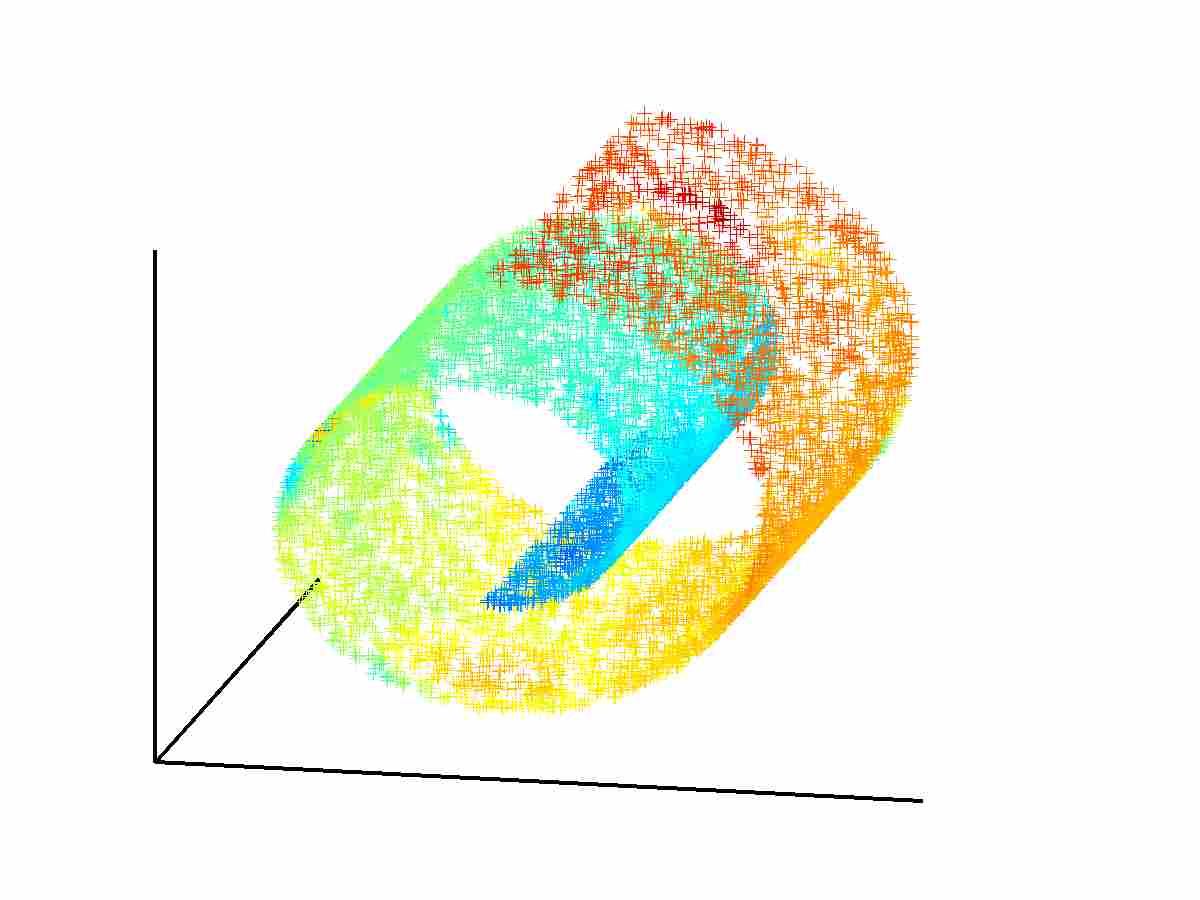} &
    \includegraphics[width=0.23\textwidth]{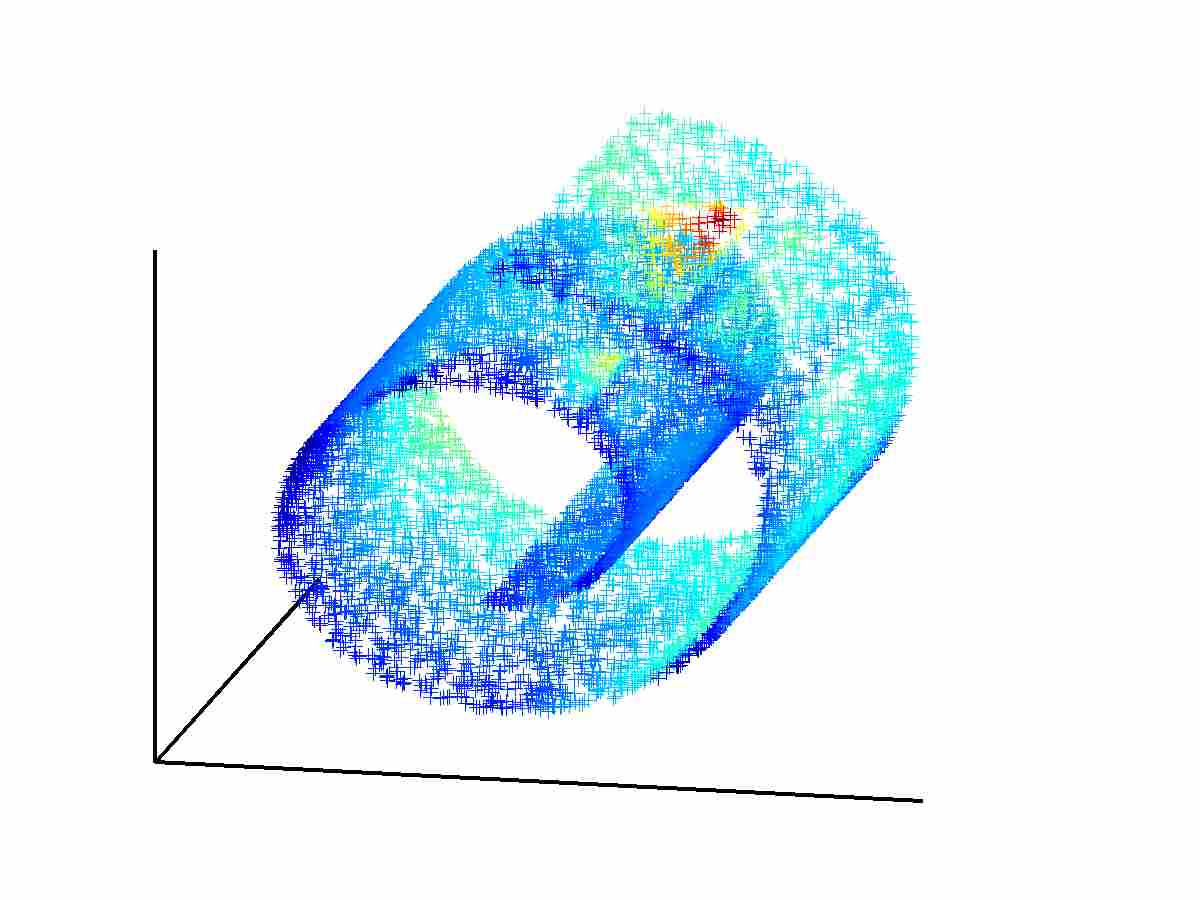}    \\
        (1) A nonlinear function & (2) RMSE=$4.394$ & (3) RMSE=$0.499$  & (4)RMSE=$4.661$  \\
    \includegraphics[width=0.23\textwidth]{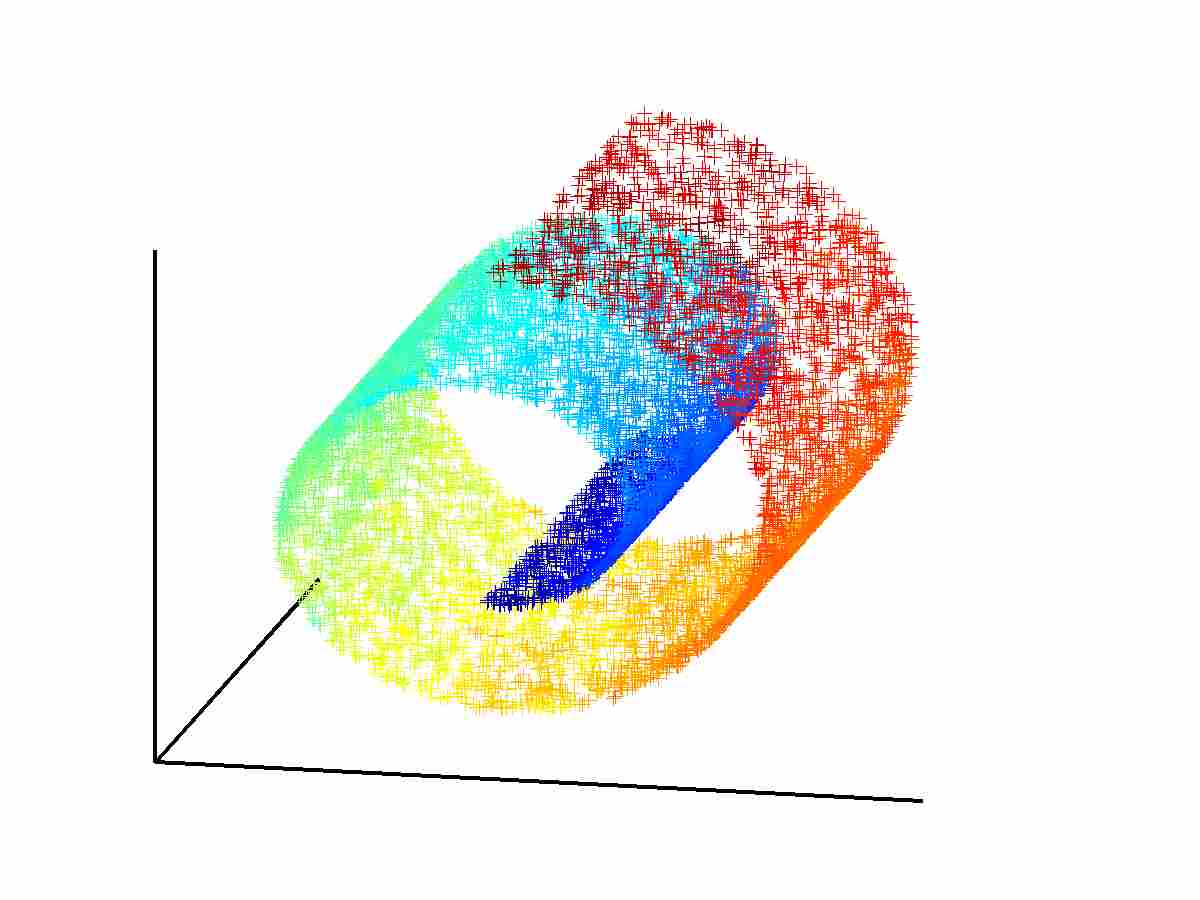} &
   \includegraphics[width=0.23\textwidth]{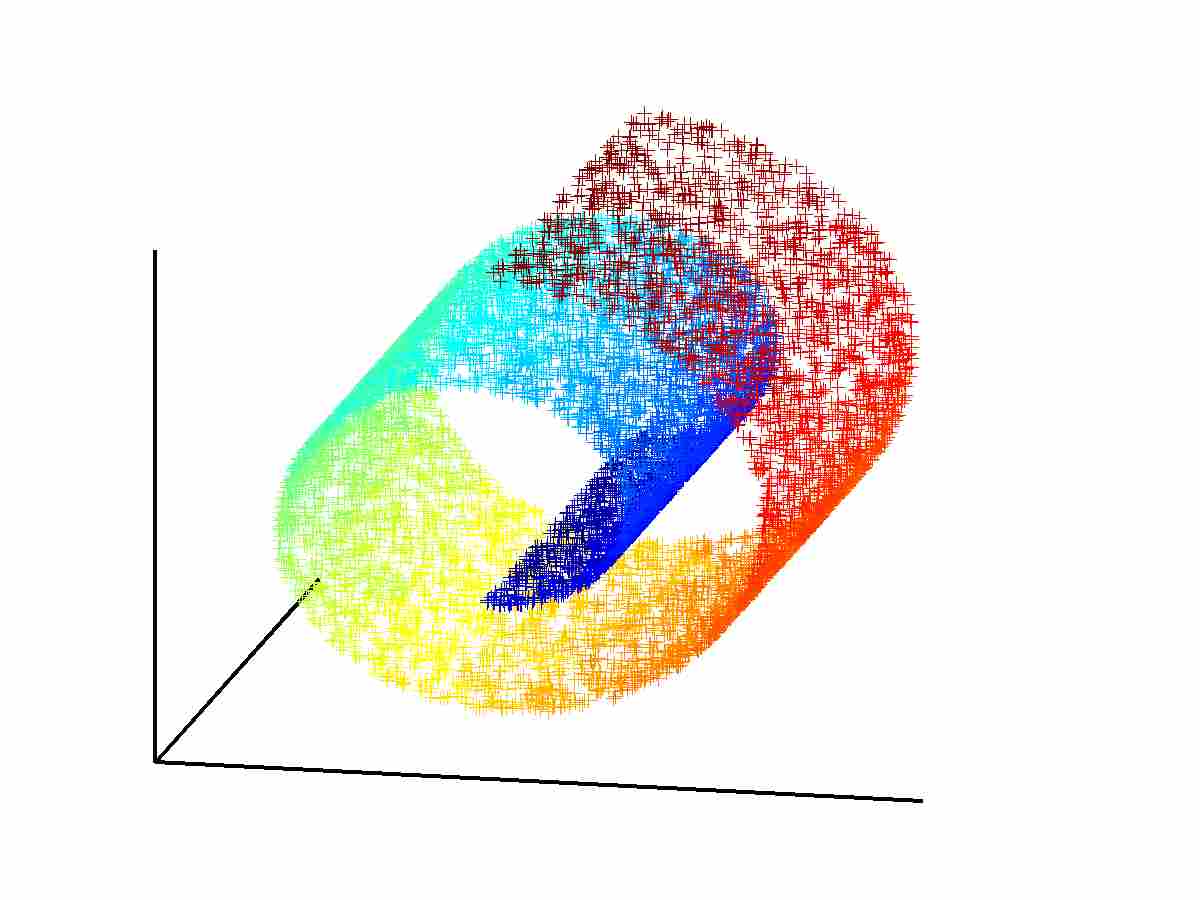} &
    \includegraphics[width=0.23\textwidth]{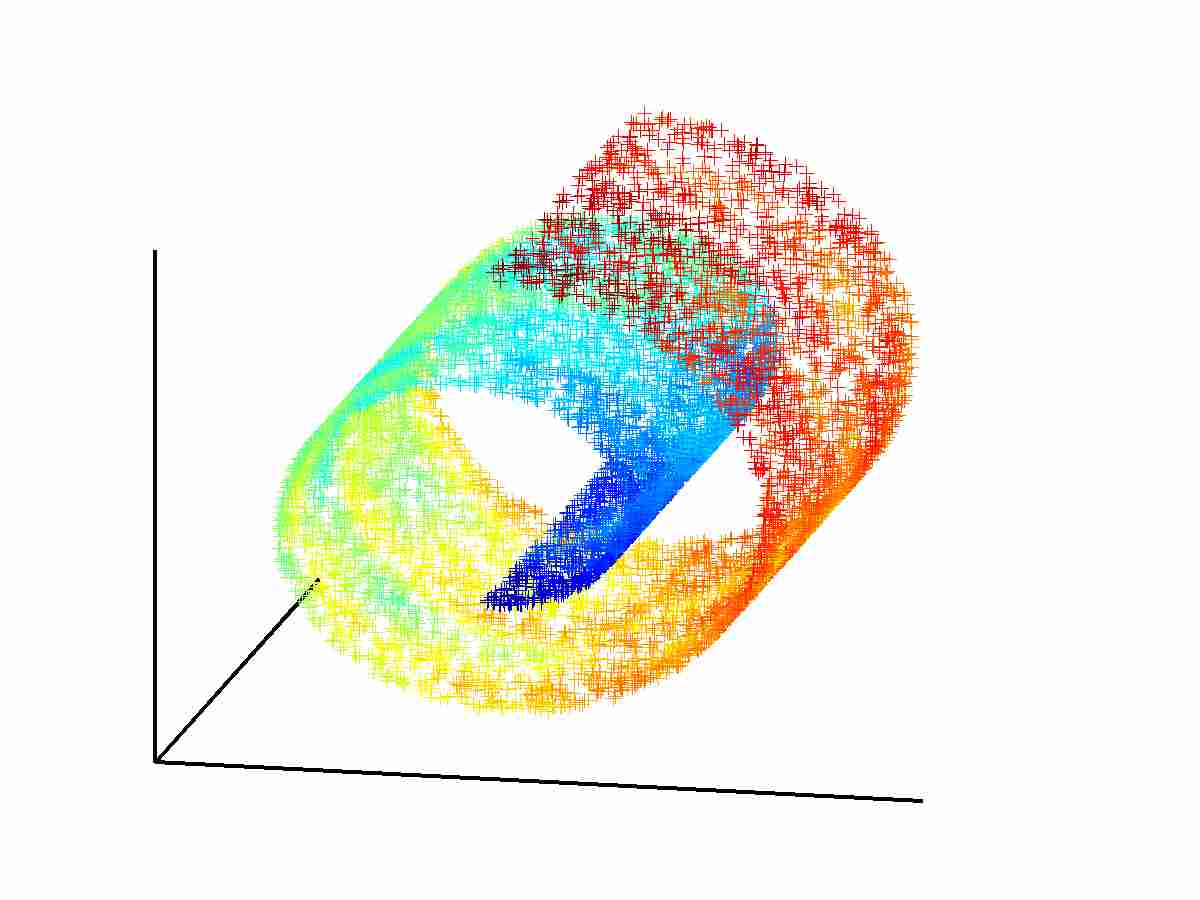} &
   \includegraphics[width=0.23\textwidth]{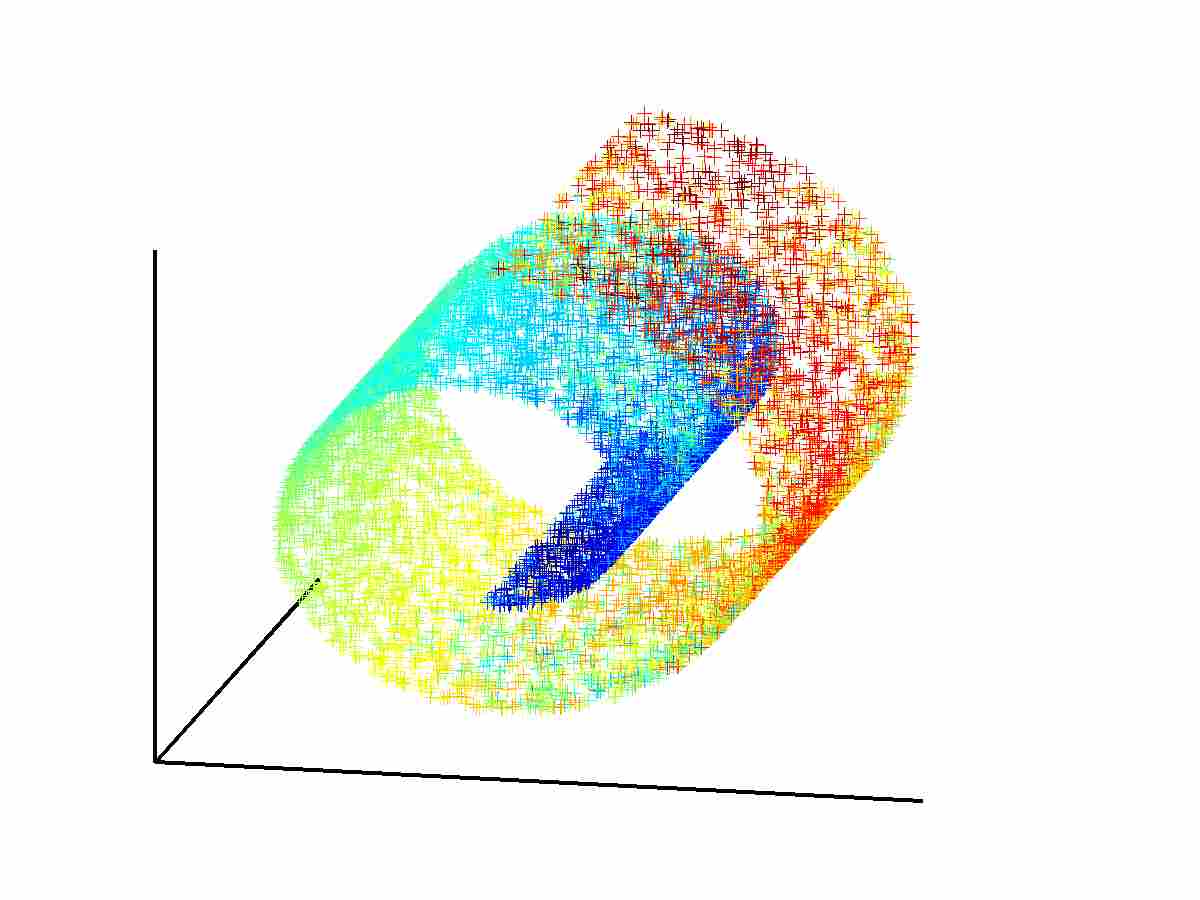}\\
       (5) RMSE=$0.201$ & (6) RMSE=$0.109$ & (7) RMSE=$0.669$ &
       (8) RMSE=$1.170$
  \end{tabular} \normalsize
  \caption{Experiments of nonlinear regression on the Swiss-roll data: (1) a nonlinear function on
  the Swiss-roll manifold, where the color indicates function values; (2) result of sparse coding with fixed random anchor points;
  (3) result of local coordinate coding with fixed random anchor points; 4) result of sparse coding; (5) result of local coordinate
  coding; (6) result of local kernel smoothing; (7) result of local coordinate coding on noisy data;
  (8) result of local kernel smoothing on noisy
  data.}
  \label{fig:swissroll}
\end{figure}

Our first experiment is based on a synthetic data set, where a
nonlinear function is defined on a Swiss-roll manifold, as shown in
Figure \ref{fig:swissroll}-(1). The primary goal is to demonstrate
the performance of nonlinear function learning using simple linear
ridge regression based on representations obtained from traditional
sparse coding and the newly suggested local coordinate coding, which
are, respectively, formulated as the following,
\begin{align} \label{eq:sparse_coding}
\min_{\gamma,C}~\sum_x \frac{1}{2}\left\|x- \gamma(x)\right\|^2
  + \beta \sum_{v \in C} |\gamma_v(x)|+ \lambda \sum_{v \in C} \|v\|^2
\end{align}
\begin{align} \label{eq:approximated_localcoding}
\min_{\gamma,C}~\sum_x\frac{1}{2}\left\|x- \gamma(x)\right\|^2
  + \beta \sum_{v \in C} |\gamma_v(x)|\|v-x\|^2 + \lambda \sum_{v \in C} \|v\|^2
\end{align}
where $\gamma(x)=\sum_{v\in C} \gamma_v(x) v$. We note that
\eqref{eq:approximated_localcoding} is an approximation to the
original formulation, mainly for the simplicity of computation.

We randomly sample $50,000$ data points on the manifold for
unsupervised basis learning, and $500$ labeled points for supervised
regression. The number of bases is fixed to be 128. The learned
nonlinear functions are tested on another set of $10,000$ data
points, with their performances evaluated by root mean square error
(RMSE).

\begin{figure}
  \centering \footnotesize
  \begin{tabular}{{c}{c}}
  \includegraphics[width=0.45\textwidth]{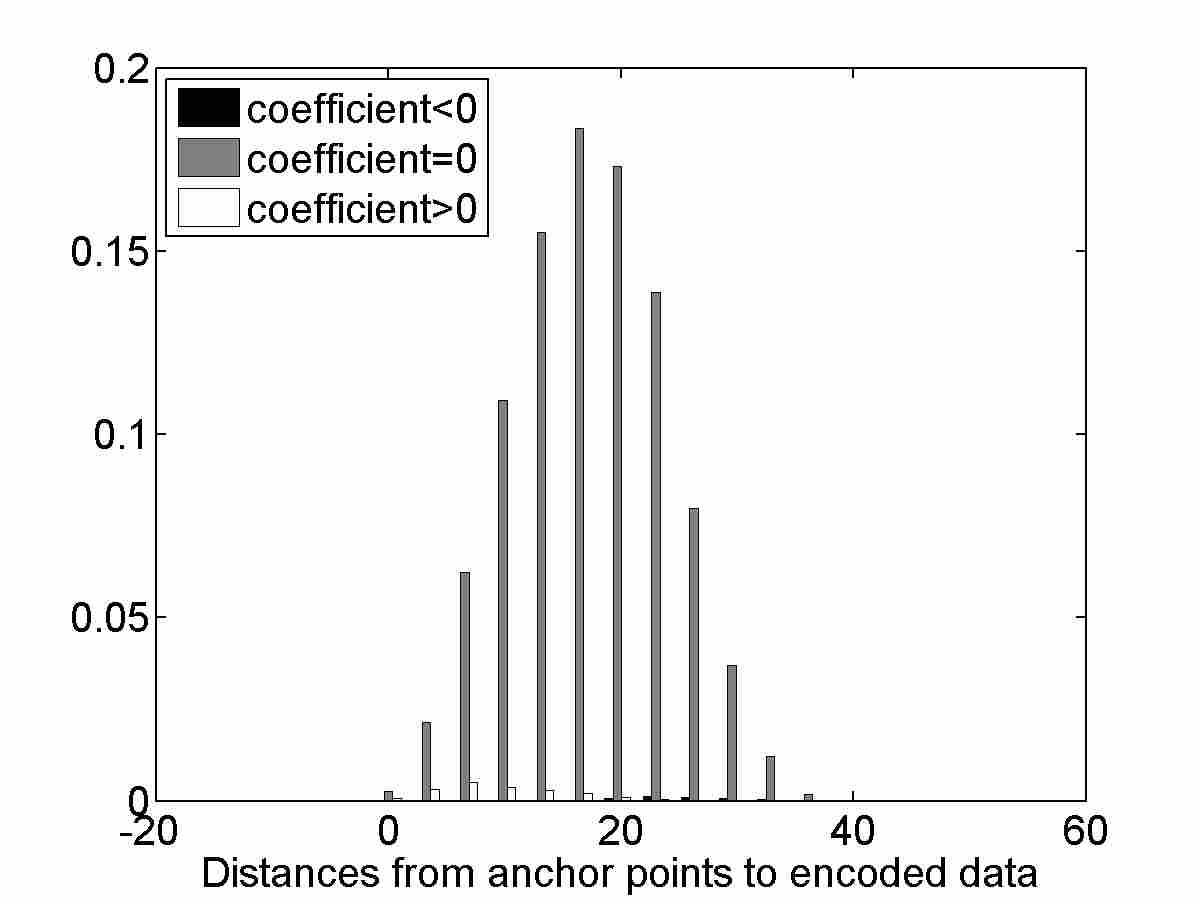} &
    \includegraphics[width=0.45\textwidth]{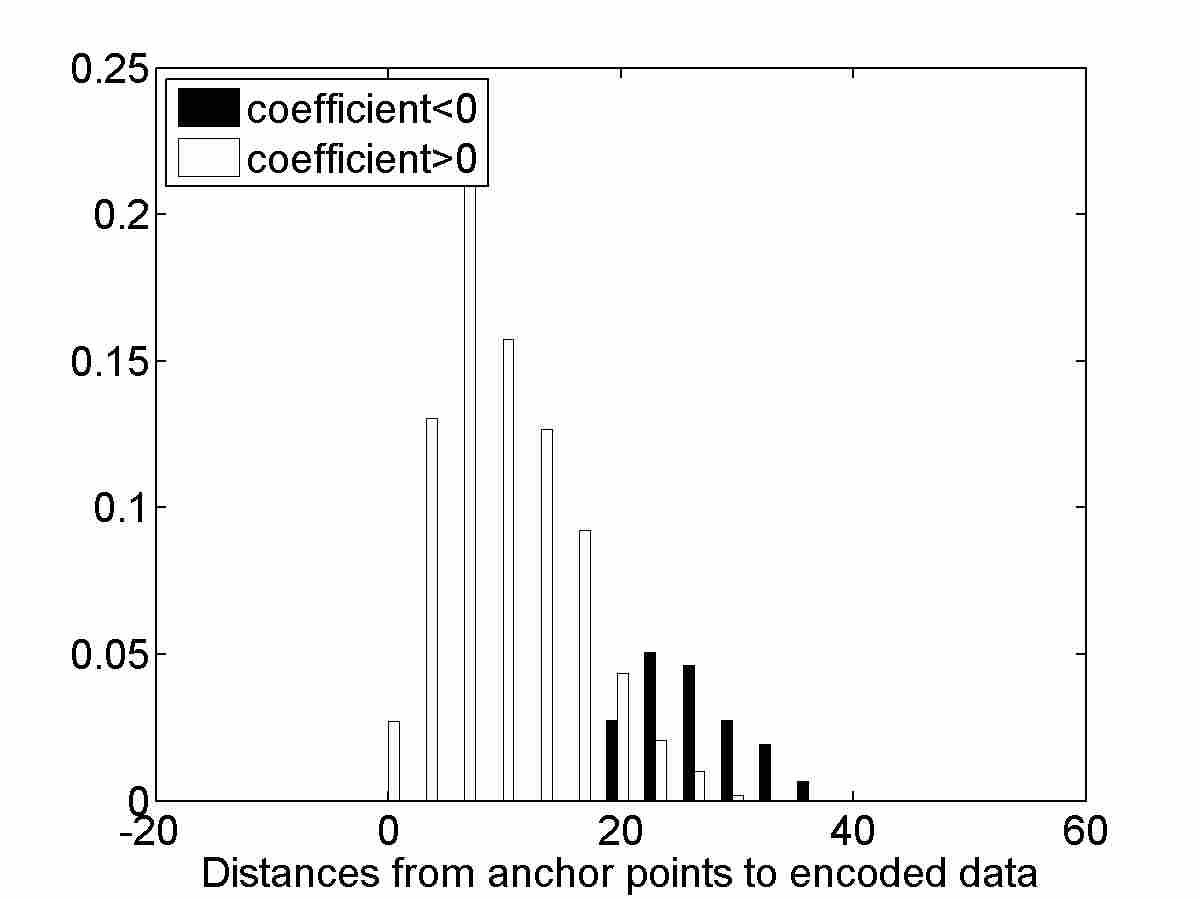} \\
     (a-1)  & (a-2)    \\
  \includegraphics[width=0.45\textwidth]{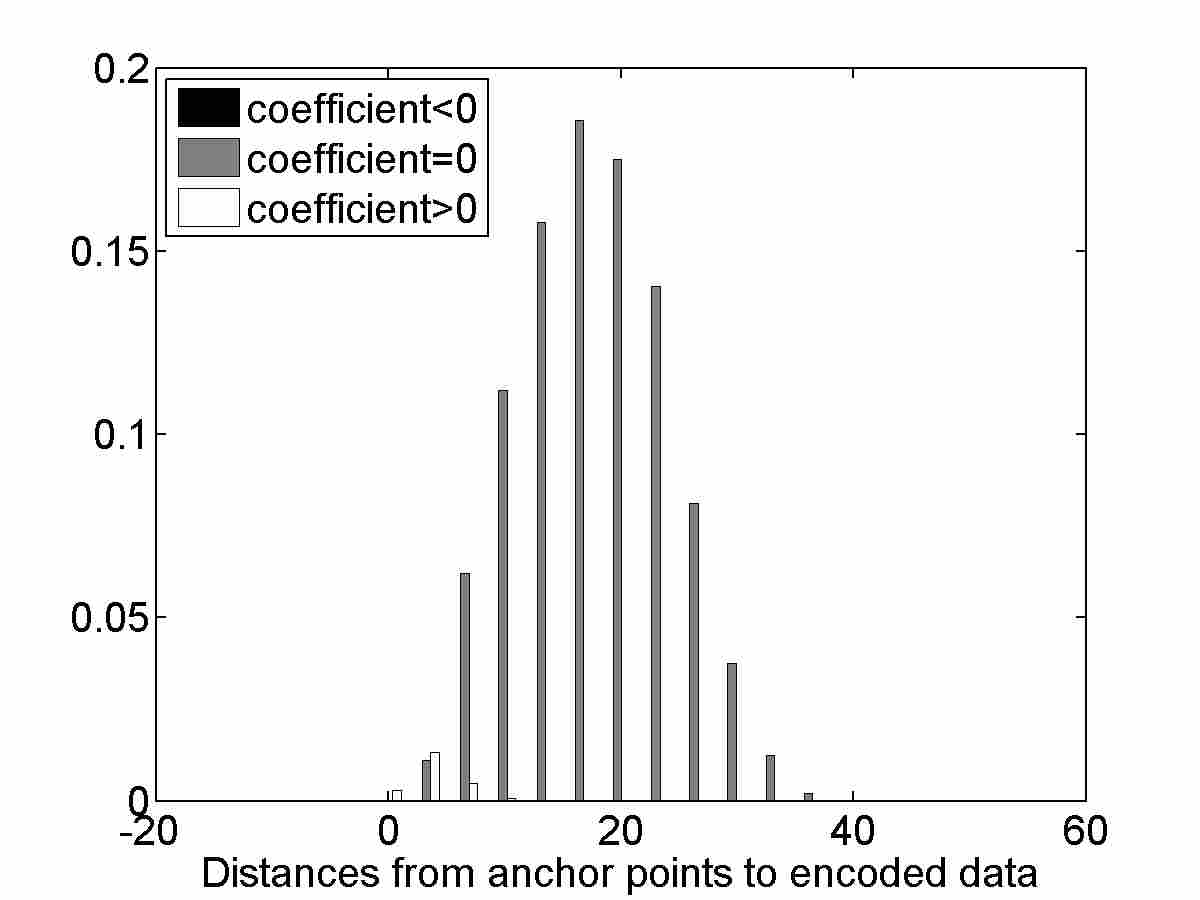} &
   \includegraphics[width=0.45\textwidth]{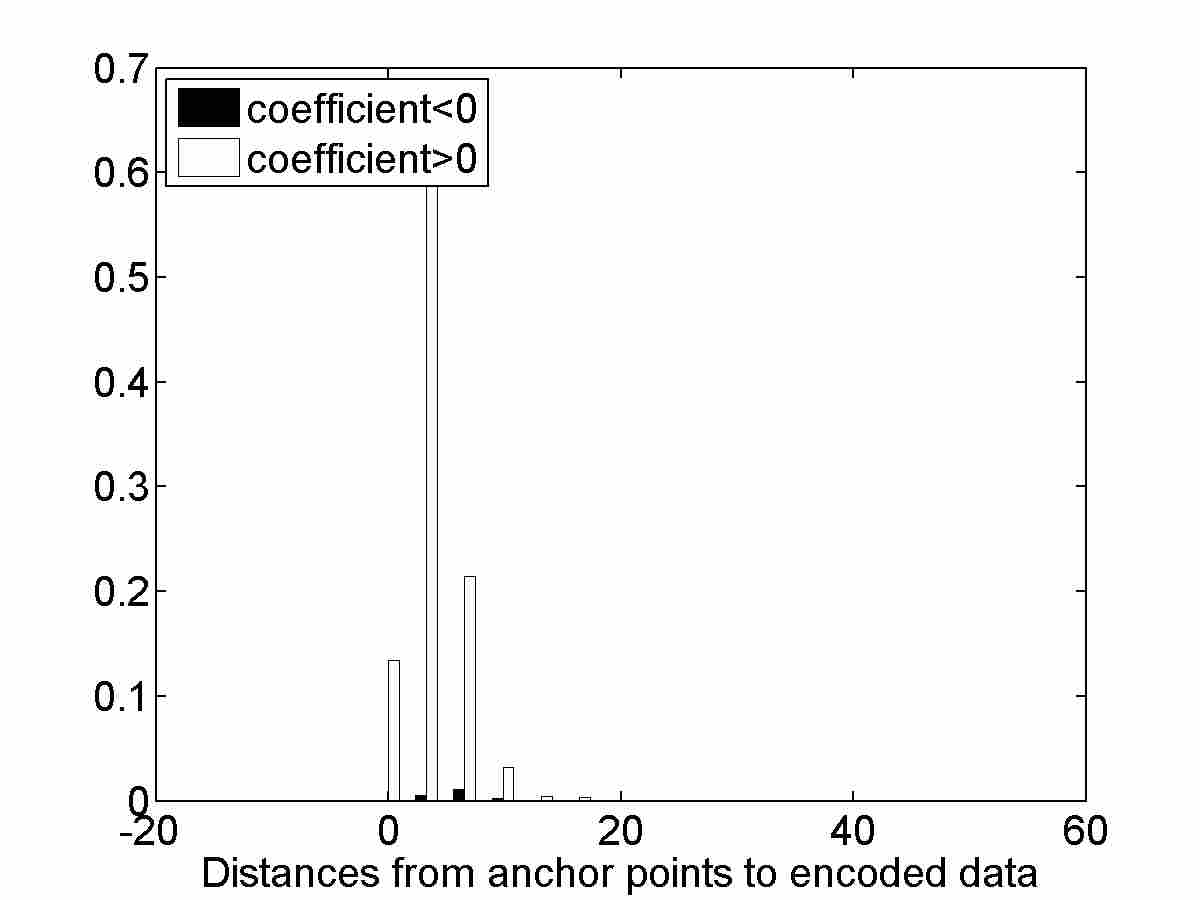}\\
  (b-1)  &
  (b-2)
  \end{tabular} \normalsize
  \caption{Coding locality on Swiss roll: (a) sparse coding  vs. (b) local coordinate coding.}
  \label{fig:swissroll_locality}
\end{figure}

In the first setting, we let both coding methods use the same set of
fixed bases, which are 128 points randomly sampled from the
manifold. The regression results are shown in
Figure~\ref{fig:swissroll}-(2) and (3), respectively. Sparse coding
based approach fails to capture the nonlinear function, while local
coordinate coding behaves much better. We take a closer look at the
data representations obtained from the two different encoding
methods, by visualizing the distributions of distances from encoded
data to bases that have positive, negative, or zero coefficients in
Figure~\ref{fig:swissroll_locality}. It shows that sparse coding
lets bases faraway from the encoded data have nonzero coefficients,
while local coordinate coding allows only nearby bases to get
nonzero coefficients. In other words, sparse coding on this data
does not ensure a good locality and thus fails to facilitate the
nonlinear function learning. As another interesting phenomenon,
local coordinate coding seems to encourage coefficients to be
nonnegative, which is intuitively understandable
--- if we use several bases close to a data point to linearly
approximate the point, each basis should have a positive
contribution. However, whether there is any merit by explicitly
enforcing non-negativity will remain an interesting future work.

In the next, given the random bases as a common initialization, we
let the two algorithms learn bases from the $50,000$ unlabeled data
points. The regression results based on the learned bases are
depicted in Figure~\ref{fig:swissroll}-(4) and (5), which indicate
that regression error is further reduced for local coordinate
coding, but remains to be high for sparse coding. We also make a
comparison with local kernel smoothing, which takes a weighted
average of function values of $K$-nearest training points to make
prediction. As shown in Figure~\ref{fig:swissroll}-(6), the method
works very well on this simple low-dimensional data, even
outperforming the local coordinate coding approach. However, if we
increase the data dimensionality to be $256$ by adding
$253$-dimensional independent Gaussian noises with zero mean and
unitary variance, local coordinate coding becomes superior to local
kernel smoothing, as shown in Figure~\ref{fig:swissroll}-(7) and
(8).
This is consistent with our theory, which suggests that
local coordinate coding can work well in high dimension; on the other hand,
local kernel smoothing is known to suffer from high dimensionality and noise.

\subsection{Handwritten Digit Recognition}

Our second experiment is based on the MNIST handwritten digit
recognition benchmark, where each data point is a $28\times 28$ gray
image, and pre-normalized into a unitary $784$-dimensional vector.
In our setting, the set $C$ of anchor points is obtained from sparse
coding, whose formulation follows \eqref{eq:sparse_coding}, with the
regularization on $v$ replaced by inequality constraints $\|v\|\leq
1$. Our focus here is not on anchor point learning, but rather on
checking whether a good nonlinear classifier can be obtained if we
enforce sparsity and locality in data representation, and then apply
simple one-against-call linear SVMs.

\begin{figure}
\begin{center}
   \includegraphics[height=8cm,width=8cm]{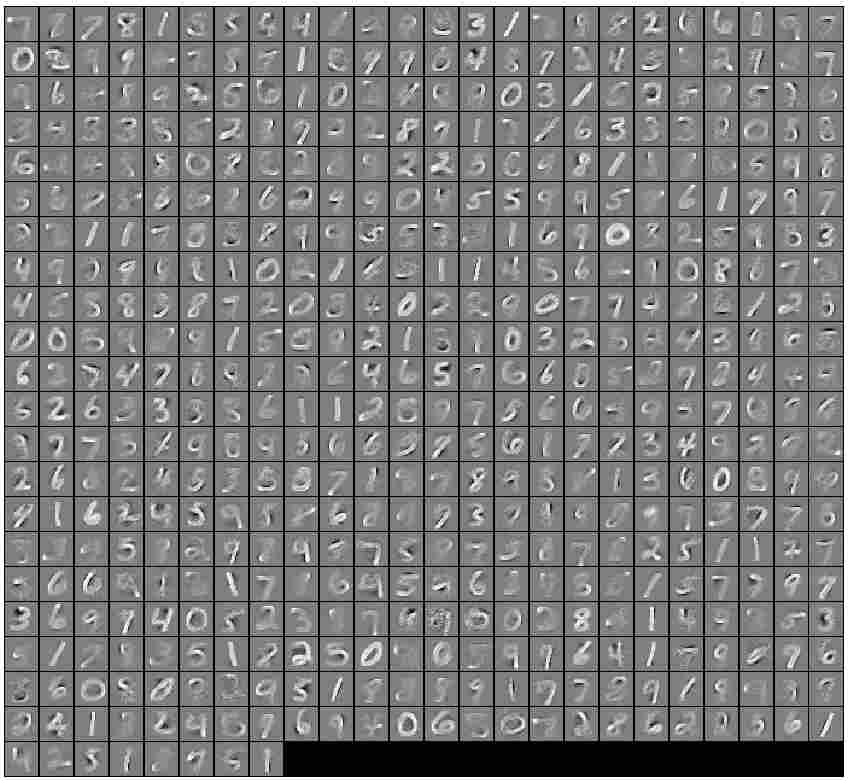}
\end{center}
\caption{The anchor points of MNIST digits ($|C|=512$).}
\label{fig:mnist_anchors_512}
\end{figure}

\begin{figure}
  \centering \footnotesize
  \begin{tabular}{{c}{c}}
  \includegraphics[width=0.45\textwidth]{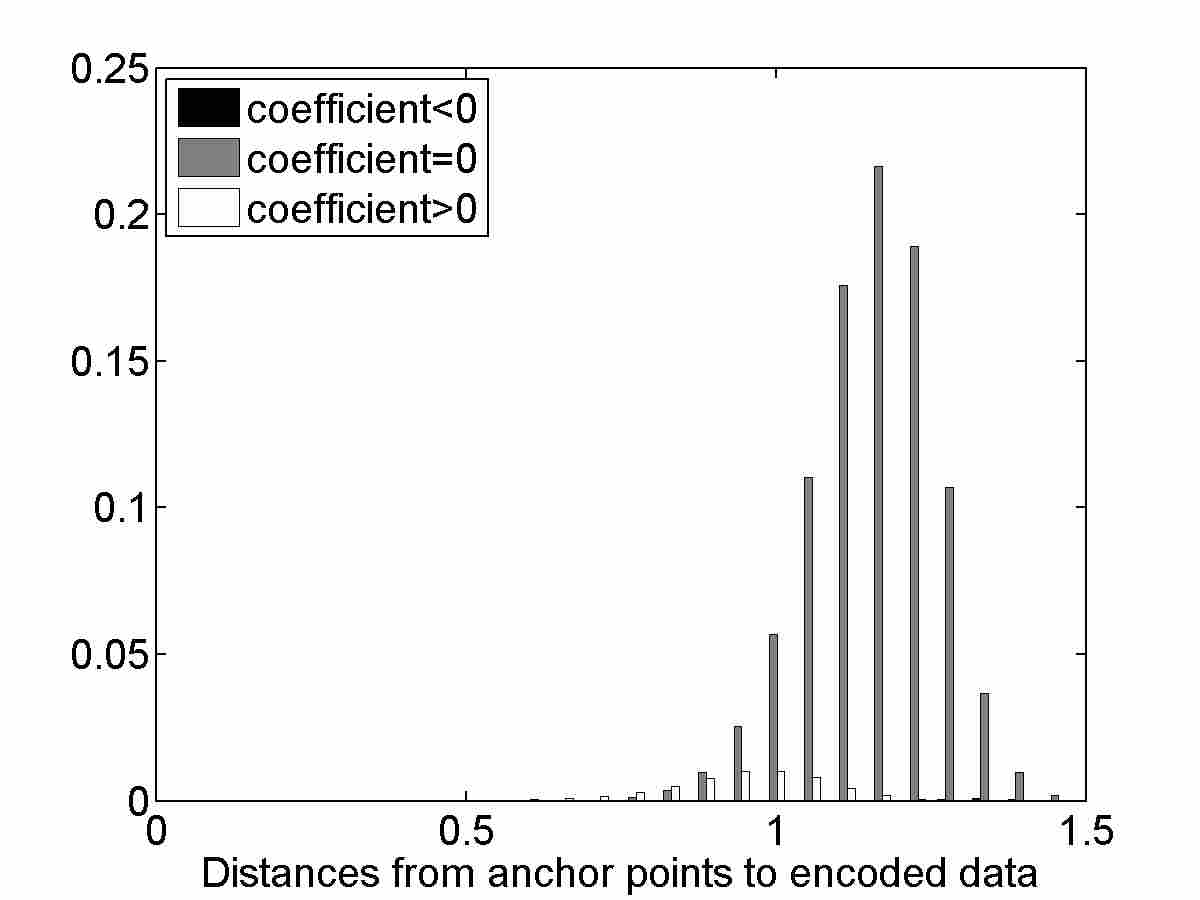} &
    \includegraphics[width=0.45\textwidth]{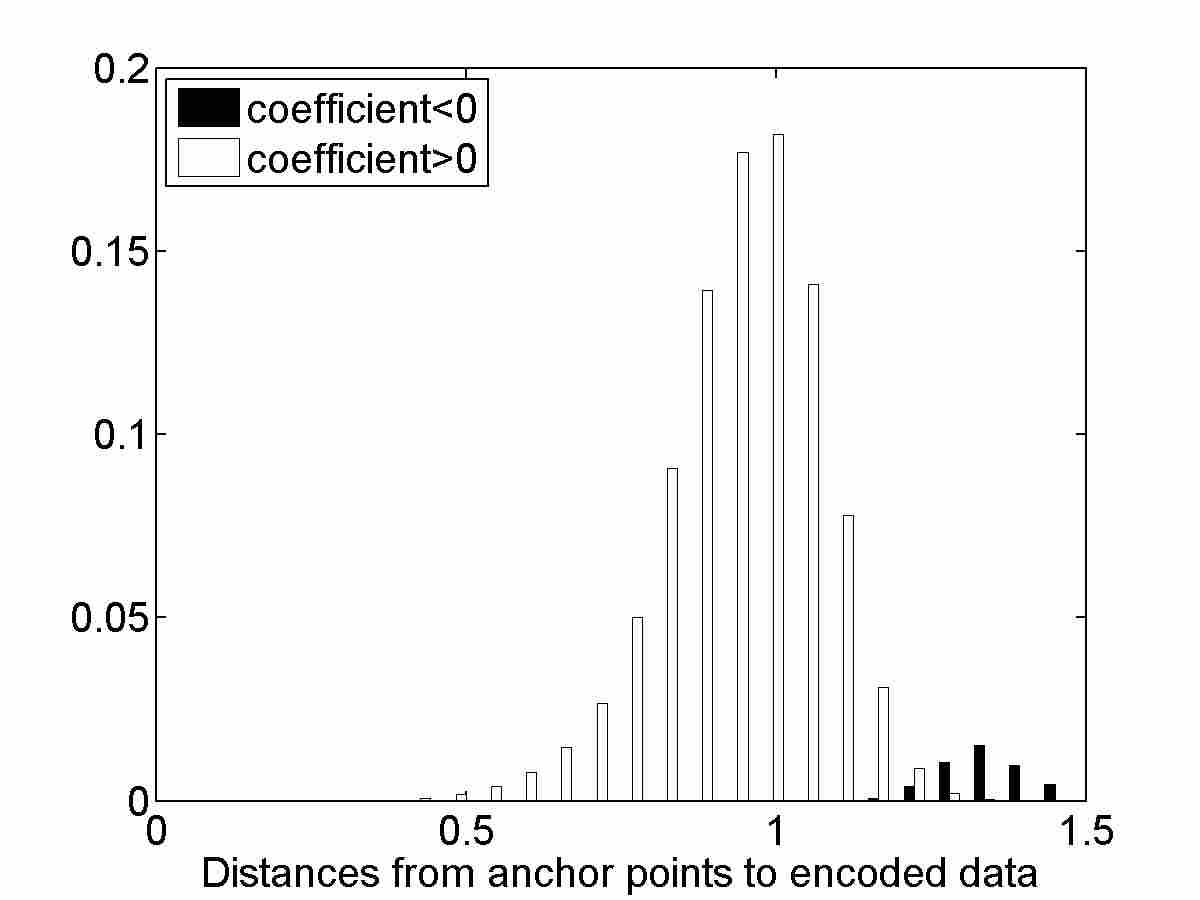} \\
     (a-1)  & (a-2)   \\
  \includegraphics[width=0.45\textwidth]{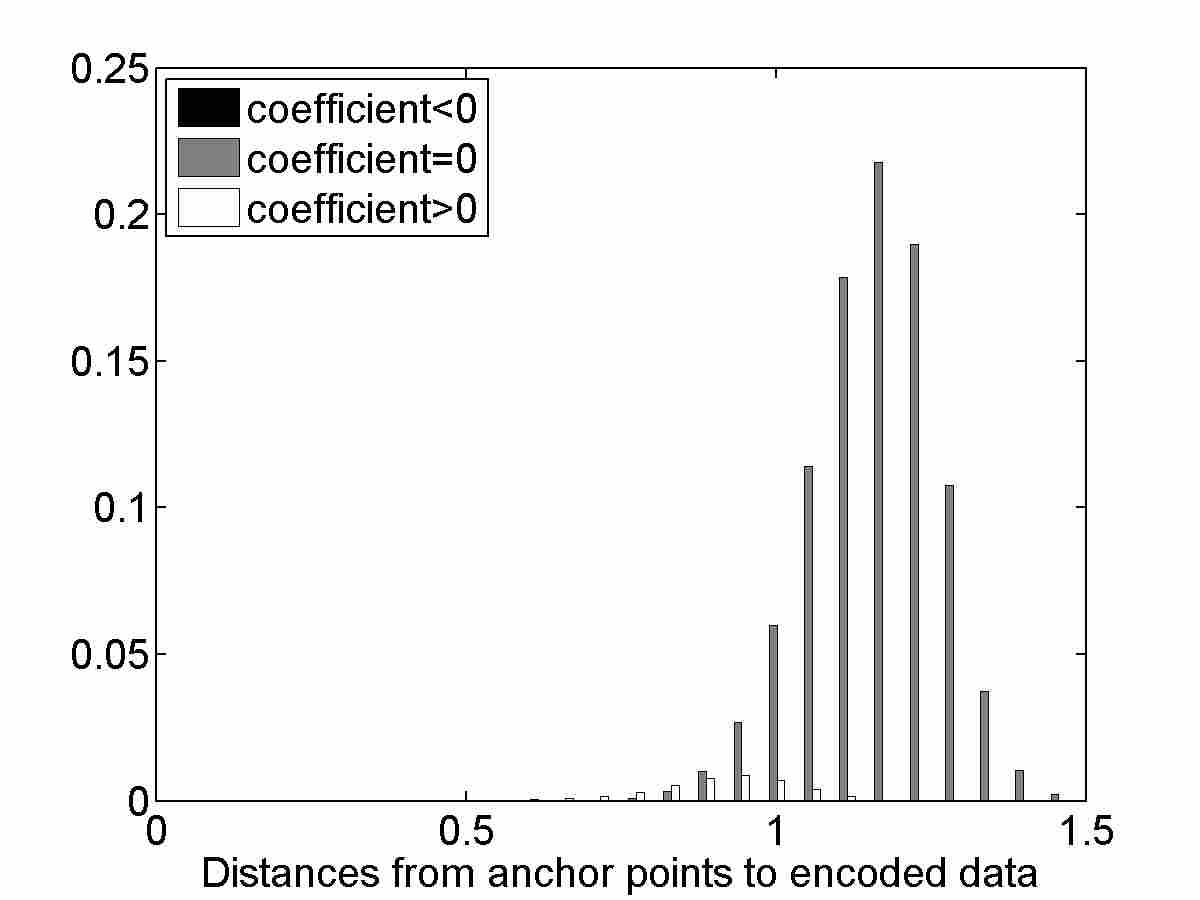} &
   \includegraphics[width=0.45\textwidth]{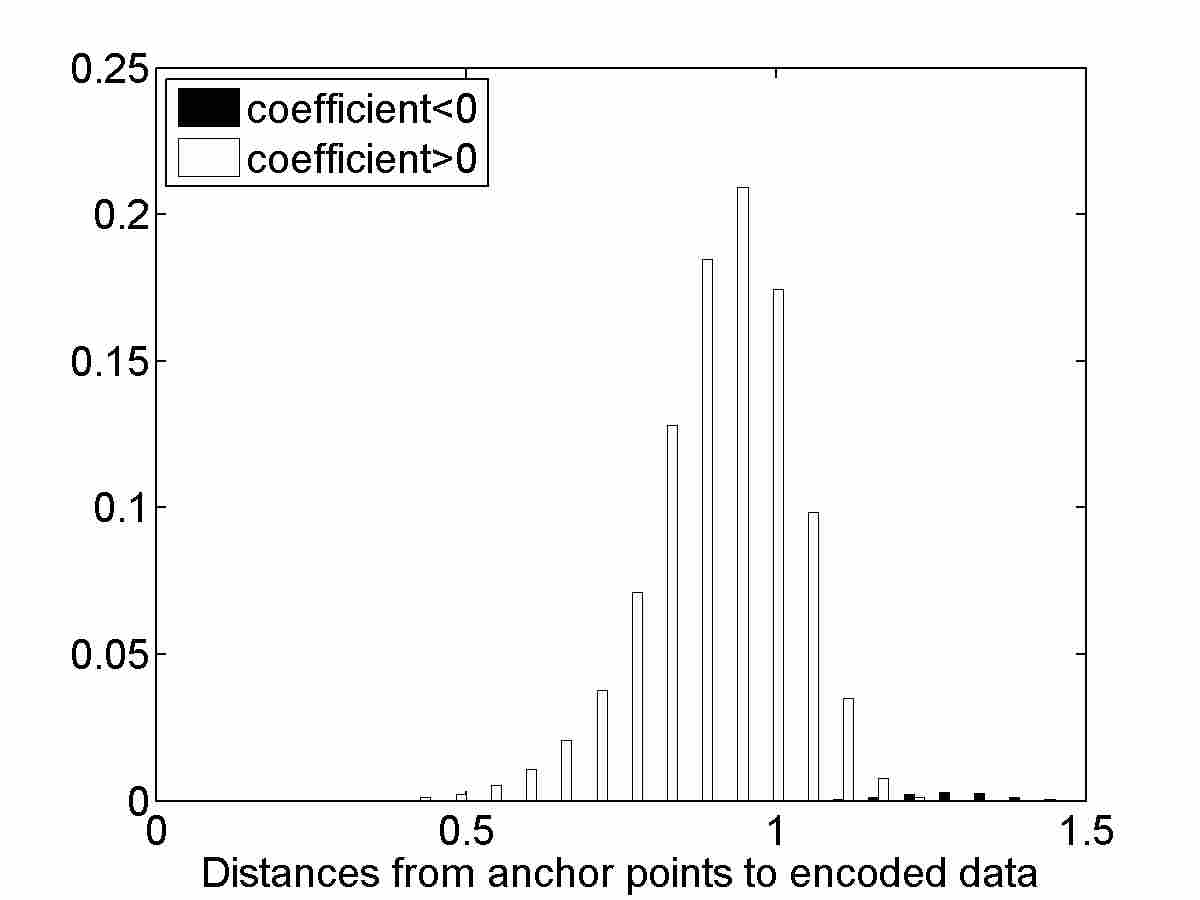}\\
  (b-1)  &
  (b-2)
  \end{tabular} \normalsize
  \caption{Coding locality on MNIST: (a) sparse coding  vs. (b) local coordinate coding. }
  \label{fig:mnist_locality}
\end{figure}

Since the optimization cost of sparse coding is invariant under
flipping the sign of $v$, we take a post-processing step to change
the sign of $v$ if we find the corresponding $\gamma_v(x)$ for most
of $x$ is negative. This rectification will ensure the anchor points
to be on the data manifold. One example of $C$ is visualized in
Figure~\ref{fig:mnist_anchors_512}, where the number of anchor
points is $|C|=512$. With the obtained $C$, for each data point $x$
we solve the local coordinate coding problem
\eqref{eq:approximated_localcoding}, by optimizing $\gamma$ only, to
obtain the representation $[\gamma_v(x)]_{v\in C}$. In the
experiments we 
try different sizes of bases.
The classification error rates are provided in
Table~\ref{tab:mnist_encoding}. In addition we also compare with
linear classifier on raw images,  local kernel smoothing based on
$K$-nearest neighbors, and linear classifiers using representations
obtained from various unsupervised learning methods, including
auto-encoder based on deep belief networks, Laplacian eigenmaps
\cite{lap_eig}, and VQ coding based on $K$-means. We note that, like
most of other manifold learning approaches, Laplacian eigenmaps is a
transductive method which has to incorporate both training and
testing data in training. The comparison results are summarized in
Table~\ref{tab:mnist_method}. Both sparse coding and local
coordinate coding perform quite good for this nonlinear
classification task, significantly outperforming linear classifiers
on raw images. In addition, local coordinate coding is consistently
better than sparse coding across various basis sizes. We further
check the locality of both representations by plotting
Figure-\ref{fig:mnist_locality}, where the basis number is $512$,
and find that sparse coding on this data set happens to be quite
local
--- unlike the case of Swiss-roll data --- here only a small portion
of nonzero coefficients (again mostly negative) are assigned onto
the bases whose distances to the encoded data exceed the average of
basis-to-datum distances. This locality explains why sparse coding
works well on MNIST data. On the other hand, local coordinate coding
is able to remove the unusual coefficients and further improve the
locality. Among those compared methods in
Table~\ref{tab:mnist_method}, we note that the error rate $1.2\%$ of
deep belief network reported in \cite{hinton06:science} was obtained
via unsupervised pre-training followed by supervised
back-propagation. The error rate based on unsupervised training of
deep belief networks is  about $1.90\%$.\footnote{This is obtained
via a personal communication with Ruslan Salakhutdinov at University
of Toronto.} 
Therefore our result is competitive to
the-state-of-the-art results that are based on unsupervised feature
learning plus linear classification
without using additional image geometric information.

\begin{table}[!ht]
\caption{Error rates (\%) of MNIST classification with different
$|C|$.}
\begin{center}
\begin{tabular}{l l l l l}
            \hline \hline
             |C| & 512 & 1024 & 2048 & 4096 \\
            \hline
            Linear SVM with sparse coding  & 2.96 & 2.64 & 2.16 &  2.02 \\
            Linear SVM with local coordinate coding & 2.64 & 2.44 & 2.08 & 1.90  \\
            \hline
\end{tabular}
\end{center} \label{tab:mnist_encoding}
\end{table}

\begin{table}[!ht]
\caption{Error rates (\%) of MNIST classification with different
methods.}
\begin{center}
\begin{tabular}{l c }
            \hline \hline
             Methods & Error Rate \\
            \hline
            Linear SVM with raw images &  12.0\\
            Local kernel smoothing & 3.48 \\
            Linear SVM with Laplacian eigenmap & 2.73 \\
            Linear SVM with VQ coding & 3.98 \\
            Linear classifier with deep belief network & 1.90 \\
            Linear SVM with sparse coding  & 2.02 \\
            Linear SVM with local coordinate coding & 1.90  \\
            \hline
\end{tabular}
\end{center} \label{tab:mnist_method}
\end{table}

\subsection{Object Recognition}

Our third experiment is image classification based on coding of
local patches. We use the the Caltech-101 benchmark, which contains
$9144$ images covering $101$ classes (including animals, vehicles,
flowers, etc.) of objects and one additional background class. The
visual patterns within each class have a high degree of variations
in translation, deformation, scale, and rotation, which requires a
data coding strategy different from that for MNIST images.
Basically, instead of coding on entire images, successful approaches
first perform coding on local patches, and then pool local codes to
obtain image-level representations. The current state-of-the-art
method \cite{SPM} on Caltech-101 takes the following steps: (1)
extraction of SIFT descriptors from local patches at a grid of
locations in an image; (2) VQ coding of each SIFT descriptor; (3)
average pooling of codes at different locations and scales; (4)
classification using a nonlinear SVM with Chi-square kernel. Here we
will examine a different method that replaces VQ coding by sparse
coding or local coordinate coding, and applies simple linear SVMs
for classification.

\begin{figure}
  \centering \footnotesize
  \begin{tabular}{{c}{c}}
  \includegraphics[width=0.45\textwidth]{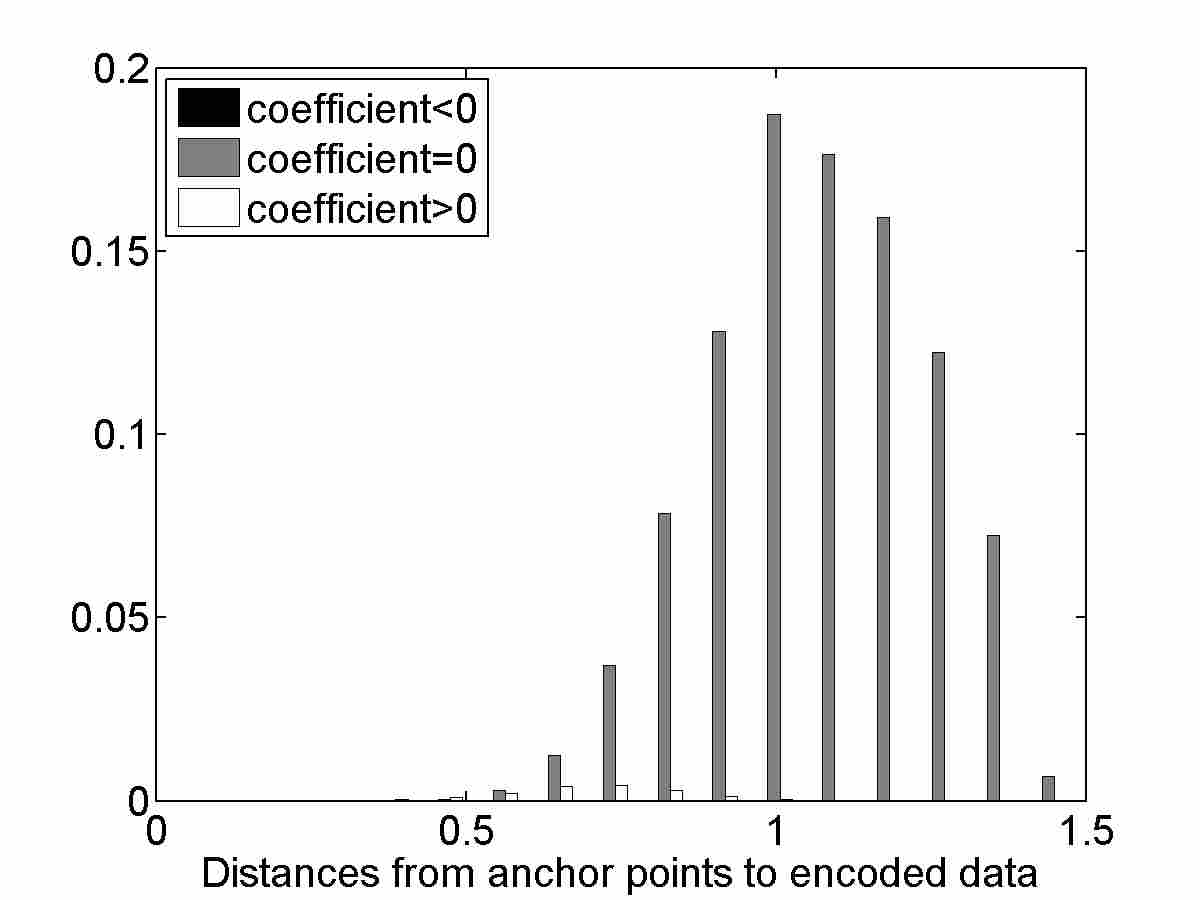} &
    \includegraphics[width=0.45\textwidth]{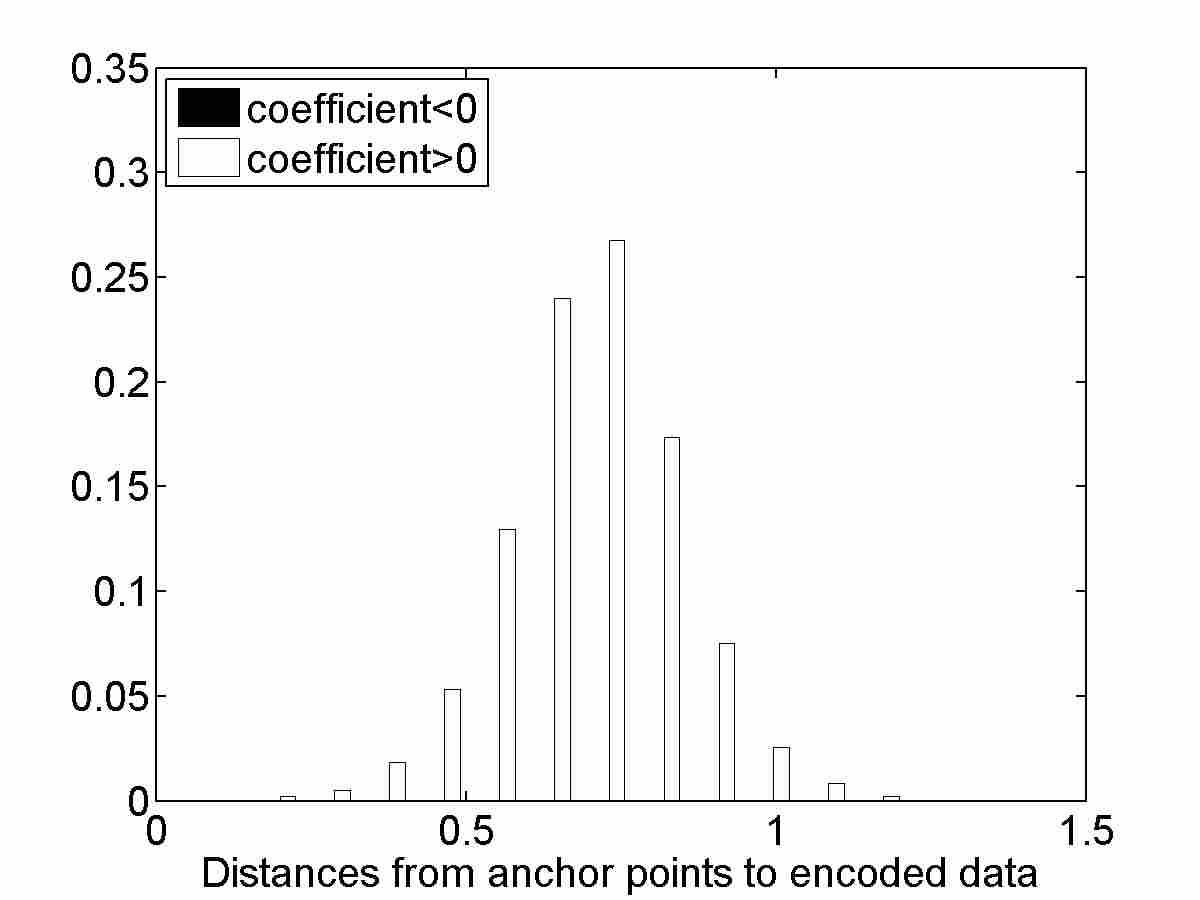} \\
     (a-1)  & (a-2)   \\
  \includegraphics[width=0.45\textwidth]{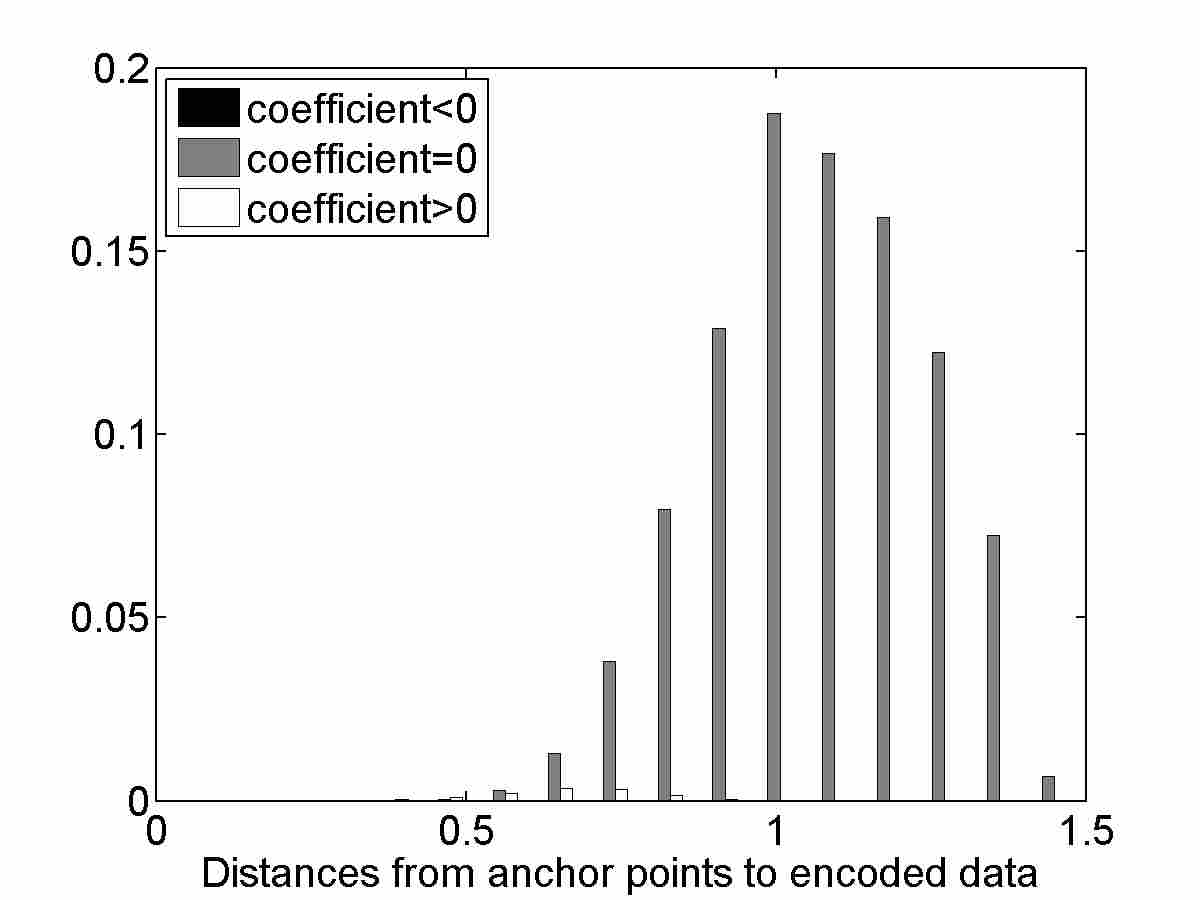} &
   \includegraphics[width=0.45\textwidth]{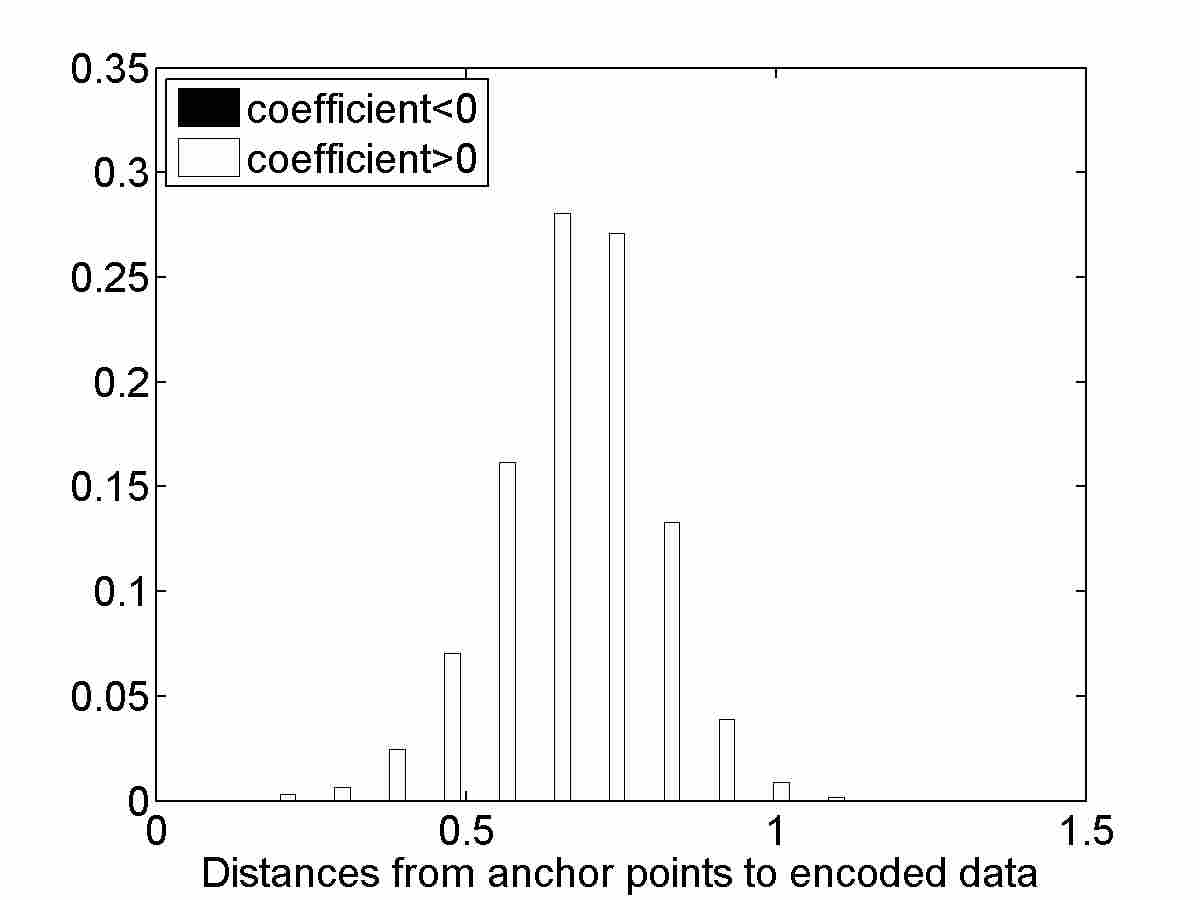}\\
  (b-1)  &
  (b-2)
  \end{tabular} \normalsize
  \caption{Coding locality on Caltech-101: (a) sparse coding  vs. (b) local coordinate coding. }
  \label{fig:caltech101_locality}
\end{figure}

We follow the common experiment setup for Caltech-101,
i.e.,~training on 30 images per category and testing on the rest,
and randomly repeat the experiments for 10 times. The step size of
the grid is $8$ pixels, and each SIFT is extracted on a $24\times
24$ patch centered at a location;  we use $200,000$ random patches'
SIFT descriptors to train the bases for VQ and sparse coding; each
image is partitioned into $1\times 1$, $2\times 2$, and $4\times 4$
blocks in 3 different scales, and pooling is done within each of the
$21$ blocks. In addition to average pooling, we also try max
pooling, i.e.,~computing the max value of each dimension of codes in
each block. Finally the pooled codes are concatenated to form a
single image-level feature vector. The results are presented in
Table~\ref{tab:caltech101}. Our methods using sparse coding and
local coordinate coding both achieve much higher accuracies on this
popular benchmark. Since only linear classifiers are required, the
methods are much more scalable and efficient for training and
testing, compared to those state-of-the-art methods relying on
nonlinear SVMs. We note that local coordinate coding does not
produce better results than sparse coding in this experiments. This
is because sparse coding in this case is already sufficiently local,
as illustrated in Figure~\ref{fig:caltech101_locality}. This result
again is consistent with the main point of the paper, that is,
coding locality is essential (and sufficient) for
ensuring a good nonlinear learning performance.

\begin{table}[!ht]
\begin{center}
\caption{Classification rate (\%) comparison on Caltech-101. }
\label{tab:caltech101}
\begin{tabular}{l c}
\hline \hline
Methods & Accuracy \\
\hline
VQ coding, average pooling, linear SVM  & $58.81\pm 1.51$ \\
VQ coding, average pooling, nonlinear SVM & $63.99\pm 0.88$ \\
Sparse coding, average pooling, linear SVM  & ${66.68\pm 0.66}$ \\
Sparse coding, max pooling, linear SVM  & ${73.20\pm 0.54}$ \\
Local coordinate coding, average pooling, linear SVM  &  ${66.72\pm 0.52}$\\
Local coordinate coding, max pooling, linear SVM  &  ${73.14\pm 0.48}$\\
\hline
\end{tabular}
\end{center}
\end{table}

\section{Conclusion}

This paper introduces a new method for high dimensional nonlinear
learning with data distributed on manifolds. 
The method can be seen
as generalized local linear function approximation, but can be
achieved by learning a global linear function with respect to
coordinates from unsupervised local coordinate coding. Compared to
popular manifold learning methods, our approach can naturally handle
unseen data and has a linear complexity with respect to data size.
The work also generalizes popular VQ coding and sparse coding
schemes, and reveals that locality of coding is essential for
supervised function learning. The generalization performance obtained
in the paper depends
on intrinsic dimensionality of the data manifold. The experiments on
synthetic data, handwritten digit data, and object-class images
further confirm the findings of our analysis.

While some results in the paper explicitly rely on the manifold concept,
the main idea is more general than manifold learning. 
The theory is valid even when the data do not
lie on a manifold. In fact, the manifold structure is only used to
bound the complexity of the local coordinate coding scheme, while
the general theory can still be applied 
if we estimate the complexity using other means.

Finally, it is worth mentioning that on many real data, sparse coding 
(without locality constraint) automatically produces coding schemes that
are nearly local. This explains the practical success of sparse coding.
It remains an interesting question to investigate conditions under which sparse codes are local, since such conditions directly 
imply the effectiveness of sparse coding according to our theory.

\section{Proofs}

\subsection{Proof of Proposition~\ref{prop:shift-inv}}
Consider a change of the $\Real^d$ origin by $u \in \Real^d$, which
shifts any point $x \in \Real^d$ to $x + u$, and points $v \in C$ to
$v+u$. The shift-invariance requirement implies that after the
change, we map $x+u$ to $\sum_{v \in C} \gamma_v(x) v + u$, which
should equal $\sum_{v \in C} \gamma_v(x) (v+u)$. This is equivalent
to $u= \sum_{v \in C} \gamma_v(x) u$, which holds if and only if
$\sum_{v \in C} \gamma_v(x)=1$.

\subsection{Proof of Lemma~\ref{lem:coding}}
For simplicity, let $\gamma_v=\gamma_v(x)$ and
  $x'=\gamma(x)=\sum_{v \in C} \gamma_v v$. We have
  \begin{align*}
    &|f(x) - \sum_{v \in C} \gamma_v f(v)| \\
    \leq&|f(x)-f(x')|+ \left|\sum_{v \in C} \gamma_v (f(v)-f(x'))\right| \\
    =&|f(x)-f(x')|+
    \left|\sum_{v \in C} \gamma_v (f(v)-f(x')- \nabla f(x')^\top(v-x'))\right| \\
    \leq&|f(x)-f(x')|+ \sum_{v \in C} |\gamma_v| |(f(v)-f(x')- \nabla f(x')^\top(v-x'))| \\
    \leq & \alpha \|x-x'\|_2 + \beta \sum_{v \in C} |\gamma_v| \|x'-v\|^{1+p} .
\end{align*}
This implies the bound.

\subsection{Proof of Theorem~\ref{thm:manifold-coding}}
Given any $\epsilon>0$, consider an $\epsilon$-cover
$C'$ of $\cM$ with $|C'| \leq \cN(\epsilon,\cM)$. Given each $u \in
C'$, define $C_u=\{v_1(u),\ldots,v_d(u)\}$, where $v_j(u)$ are
defined in Definition~\ref{def:manifold}. Define the anchor points
as
\[
C= \cup_{u \in C'} \{u+v_j(u): j=1,\ldots,m\}  \cup C' .
\]
It follows that $|C| \leq (1+m) \cN(\epsilon,\cM)$.

In the following, we only need to prove the existence of a coding
$\gamma$ on $\cM$ that satisfies the requirement of the theorem.
Without loss of generality, we assume that $\|v_j(u)\|=\epsilon$ for
each $u$ and $j$, and given $u$, $\{v_j(u): j=1,\ldots,m\}$ are
orthogonal with respect to $A$: $v_j^\top(u) A v_k(u)=0$ when $j \neq
k$.

For each $x \in \cM$, let $u_x \in C'$ be the closest point to $x$
in $C'$. We have $\|x-u_x\| \leq \epsilon$ by the definition of
$C'$. Now, Definition~\ref{def:manifold} implies that there exists
$\gamma_j'(x)$ ($j=1,\ldots, m$) such that
\[
\left\|x -u_x - \sum_{j=1}^m \gamma_j'(x) v_j(u_x)\right\| \leq
c_p(\cM) \epsilon^{1+p} .
\]
The optimal choice is the $A$-projection of $x-u_x$ to the subspace
spanned by $\{v_j(u_x): j=1,\ldots,m\}$. The orthogonality condition
thus implies that
\[
\sum_{j=1}^m \gamma_j'(x)^2 \|v_j(u_x)\|^2 \leq \|x-u_x\|^2 \leq
\epsilon^2 .
\]
Therefore
\[
\sum_{j=1}^m \gamma_j'(x)^2 \leq 1 ,
\]
which implies that for all $x$:
\[
\sum_{j=1}^m |\gamma_j'(x)| \leq \sqrt{m} .
\]

We can now define the coordinate coding of $x \in \cM$ as
\[
\gamma_v(x)= \left\{
\begin{array}{cc}
\gamma_j'(x) & v=u_x + v_j(u_x) \\
1- \sum_{j=1}^m \gamma_j'(x)  & v=u_x\\
0         & \text{otherwise}
\end{array}
\right. .
\]
This implies the following bounds:
\[
\|x-\gamma(x)\| \leq c_p(\cM) \epsilon^{1+p}
\]
and
\begin{align*}
&\sum_{v \in C} \gamma_v(x) \left\|v-
\gamma(x)\right\|^{1+p} \\
=&|\gamma_{u_x}(x)| \|\gamma(x)-u_x\|^{1+p}
+ \sum_{j=1}^m |\gamma_j'(x)|  \left\|v_j(u_x)- (\gamma(x)-u_x)\right\|^{1+p}\\
\leq& (1+\sqrt{m}) \epsilon^{1+p} +
\sum_{j=1}^m |\gamma_j'(x)|  (\epsilon + \epsilon)^{1+p}\\
=& [1+ \sqrt{m} + 2^{1+p} \sqrt{m}] \epsilon^{1+p} ,
\end{align*}
where we have used $\|v-u_x\|=\epsilon$, and 
$\|\gamma(x)-u_x\| \leq \|x-u_x\| \leq \epsilon$
(note that $\gamma(x)-u_x$ is the projection of $x-u_x$).

\subsection{Proof of Theorem~\ref{thm:gen}}
Consider $n+1$ samples
$S_{n+1}=\{(x_1,y_1),\ldots,(x_{n+1},y_{n+1})\}$. We shall introduce
the following notation:
\begin{equation}
[\tilde{w}_v] = \arg\min_{[w_v]} \left[ \frac{1}{n} \sum_{i=1}^{n+1}
\phi\left(f_{\gamma,C}(w, x_i),y_i\right) + \lambda \sum_{v \in C}
w_v^2 \right] . \label{eq:proof-loo-tildew}
\end{equation}
Let $k$ be an integer randomly drawn from $\{1,\ldots,n+1\}$. Let
$[\hat{w}^{(k)}_v]$ be the solution of
\[
[\hat{w}_v^{(k)}] = \arg\min_{[w_v]} \left[ \frac{1}{n}
\sum_{i=1,\ldots, n+1; i \neq k} \phi\left(f_{\gamma,C}(w,
x_i),y_i\right) + \lambda \sum_{v \in C} w_v^2 \right] ,
\]
with the $k$-th example left-out.

We have the following stability lemma from  \cite{Zhang02-loo},
which can be stated as follows using our terminology:
\begin{lemma}
  The following inequality holds
  \[
  |f_{\gamma,C}(\hat{w}^{(k)},x_k)-f_{\gamma,C}(\tilde{w},x_k)|
  \leq \frac{\|x_k\|_\gamma^2}{2 \lambda n} |\phi_1'(f_{\gamma,C}(\tilde{w},x_k), y_k)| .
  \]
  \label{lem:loo}
\end{lemma}

By using Lemma~\ref{lem:loo}, we obtain for all $\alpha>0$:
\begin{align*}
&\phi(f_{\gamma,C}(\tilde{w},x_k), y_k)
- \phi(f_{\gamma,C}(\hat{w}^{(k)},x_k), y_k) \\
=& \phi(f_{\gamma,C}(\tilde{w},x_k), y_k) -
\phi(f_{\gamma,C}(\hat{w}^{(k)},x_k), y_k) \\
&\quad - \phi_1'(f_{\gamma,C}(\hat{w}^{(k)},x_k), y_k) (f_{\gamma,C}(\tilde{w},x_k)-f_{\gamma,C}(\hat{w}^{(k)},x_k)) \\
&\quad + \phi_1'(f_{\gamma,C}(\hat{w}^{(k)},x_k), y_k)
(f_{\gamma,C}(\tilde{w},x_k)-f_{\gamma,C}(\hat{w}^{(k)},x_k))
\\
\geq& \phi_1'(f_{\gamma,C}(\hat{w}^{(k)},x_k), y_k) (f_{\gamma,C}(\tilde{w},x_k)-f_{\gamma,C}(\hat{w}^{(k)},x_k))\\
\geq& -\phi_1'(f_{\gamma,C}(\hat{w}^{(k)},x_k), y_k)^2
  \|x_k\|_\gamma^2 / (2 \lambda n) \\
\geq& -B^2  \|x_k\|_\gamma^2/ (2 \lambda n) .
\end{align*}
In the above derivation, the first inequality uses the convexity of
$\phi(f,y)$ with respect to $f$, which implies that
$\phi(f_1,y)-\phi(f_2,y)-\phi_1'(f_2,y)(f_1-f_2) \geq 0$. The second
inequality uses Lemma~\ref{lem:loo}, and the third inequality uses
the assumption of the loss function.

Now by summing over $k$, and consider any fixed $f \in
\cF_{\alpha,\beta,p}$, we obtain:
\begin{align*}
& \sum_{k=1}^{n+1} \phi(f_{\gamma,C}(\hat{w}^{(k)},x_k), y_k) \\
\leq& \sum_{k=1}^{n+1} \left[ \phi(f_{\gamma,C}(\tilde{w},x_k), y_k)
+ \frac{B^2}{2 \lambda n} \|x_k\|_\gamma^2
\right] \\
\leq& n \left[ \frac{1}{n} \sum_{k=1}^{n+1} \phi\left(\sum_{v \in C}
\gamma_v(x_k) f(v),y_k\right) + \lambda \sum_{v \in C} f(v)^2
\right] +
\frac{B^2}{2 \lambda n}  \sum_{k=1}^{n+1} \|x_k\|_\gamma^2 \\
\leq& n \left[ \frac{1}{n} \sum_{k=1}^{n+1} [
\phi\left(f(x_k),y_k\right) + B Q(x_k) ] + \lambda \sum_{v \in C}
f(v)^2 \right] + \frac{B^2}{2 \lambda n}  \sum_{k=1}^{n+1}
\|x_k\|_\gamma^2 ,
\end{align*}
where $Q(x)=\alpha \left\|x- \gamma(x)\right\|
  + \beta \sum_{v \in C} |\gamma_v(x)|  \left\|v- \gamma(x)\right\|^{1+p}$.
In the above derivation, the second inequality follows from the
definition of $\tilde{w}$ as the minimizer of
(\ref{eq:proof-loo-tildew}). The third inequality follows from
Lemma~\ref{lem:coding}. Now by taking expectation with respect to
$S_{n+1}$, we obtain
\begin{align*}
&(n+1) \rE_{S_{n+1}} \phi(f_{\gamma,C}(\hat{w}^{(n+1)},x_{n+1}), y_{n+1}) \\
\leq& n \left[ \frac{n+1}{n} \rE_{x,y} \phi\left(f(x),y\right) +
\frac{n+1}{n} B Q_{\alpha,\beta,p}(\gamma,C) + \lambda \sum_{v \in
C} f(v)^2 \right] + \frac{B^2 (n+1)}{2 \lambda n} \rE_x
\|x\|_\gamma^2 .
\end{align*}
This implies the desired bound.

\subsection{Proof of Theorem~\ref{thm:consistency}}
Note that any measurable function $f: \cM \to R$ can be approximated
by $\cF_{\alpha,\beta,p}$ with $\alpha,\beta \to \infty$ and $p=0$.
Therefore we only need to show
\[
\lim_{n \to \infty} \rE_{S_n} \; \rE_{x,y}
\phi(f_{\gamma,C}(\hat{w},x),y) =\lim_{n \to \infty} \inf_{f \in
\cF_{\alpha,\beta,p}} \rE_{x,y} \phi\left(f(x),y\right) .
\]

Theorem~\ref{thm:manifold-coding} implies that it is possible to
pick $(\gamma,C)$ such that $|C|/n \to 0$ and
$Q_{\alpha,\beta,p}(\gamma,C) \to 0$. Moreover, $\|x\|_\gamma$ is
bounded.

Given any $f \in \cF_{\alpha,\beta,0}$ and any constant
$A>0$ that is independent of $n$; 
if we let $f_A(x)=\max(\min(f(x),A),-A)$, then it is clear
that $f_A(x) \in  \cF_{\alpha,\alpha+\beta,0}$. Therefore
Theorem~\ref{thm:gen} implies that as $n \to \infty$,
\[
\rE_{S_n} \; \rE_{x,y} \phi(f_{\gamma,C}(\hat{w},x),y) \leq
\rE_{x,y} \phi\left(f_A(x),y\right) + o(1) .
\]
Since $A$ is arbitrary, we let $A \to \infty$ to obtain the desired
result.

\bibliographystyle{plain}
\bibliography{manifold}

\end{document}